\documentclass[sn-mathphys,Numbered]{sn-jnl}

\usepackage{graphicx}%
\usepackage{multirow}%
\usepackage{amsmath,amssymb,amsfonts}%
\usepackage{tabularray}
\usepackage{amssymb}
\usepackage{amsthm}%
\usepackage{mathrsfs}%
\usepackage[title]{appendix}%
\usepackage{xcolor}%
\usepackage{manyfoot}%
\usepackage{booktabs}%
\usepackage{algorithm}%
\usepackage{algorithmicx}%
\usepackage{algpseudocode}%
\usepackage{listings}%
\usepackage{subcaption}
\usepackage{array}    

\begin{document}

\title[Article Title]{A Survey of Explainable Knowledge Tracing}


\author[1]{\fnm{} \sur{Yanhong Bai}}\email{Lucky\_Baiyh@stu.ecnu.edu.cn}
\author*[1]{\fnm{} \sur{Jiabao Zhao}}\email{jbzhao@mail.ecnu.edu.cn}
\author[1]{\fnm{} \sur{Tingjiang Wei}}\email{mxdlzg@163.com}
\author[2]{\fnm{} \sur{Qing Cai}}\email{qcai@psy.ecnu.edu.cn}
\author[1]{\fnm{} \sur{Liang He}}\email{lhe@cs.ecnu.edu.cn}

\affil*[1]{\orgdiv{The School of Computer Science and Technology}, \orgname{East China Normal University}, \city{Shanghai}, \postcode{200062}, \state{Shanghai}, \country{China}}

\affil*[2]{\orgdiv{The School of Psychology and Cognitive Science}, \orgname{East China Normal University}, \city{Shanghai}, \postcode{200062}, \state{Shanghai}, \country{China}}

\abstract{With the long-term accumulation of high-quality educational data, artificial intelligence (AI) has shown excellent performance in knowledge tracing (KT). However, due to the lack of interpretability and transparency of some algorithms, this approach will result in reduced stakeholder trust and a decreased acceptance of intelligent decisions. Therefore, algorithms need to achieve high accuracy, and users need to understand the internal operating mechanism and provide reliable explanations for decisions. This paper thoroughly analyzes the interpretability of KT algorithms. First, the concepts and common methods of explainable  artificial intelligence (xAI) and knowledge tracing are introduced. Next, explainable knowledge tracing (xKT) models are classified into two categories: transparent models and “black box” models. Then, the interpretable methods used are reviewed from three stages: ante-hoc interpretable methods, post-hoc interpretable methods, and other dimensions. It is worth noting that current evaluation methods for xKT are lacking. Hence, contrast and deletion experiments are conducted to explain the prediction results of the deep knowledge tracing model on the ASSISTment2009 by using three xAI methods. Moreover, this paper offers some insights into evaluation methods from the perspective of educational stakeholders. This paper provides a detailed and comprehensive review of the research on explainable knowledge tracing, aiming to offer some basis and inspiration for researchers interested in the interpretability of knowledge tracing.}




\maketitle
\section{Introduction}\label{sec1}
The emergence and application of numerous educational tools, such as profiling and prediction \cite{ouyang2022artificial}, intelligent tutoring systems \cite{mousavinasab2021intelligent,xu2023improving}, assessment and evaluation \cite{xu2022online}, adaptive systems and personalization \cite{xiao2023personalized,ahmad2023abine}, are transforming traditional methods of teaching and learning. Knowledge tracing (KT) is an important research direction in the field of artificial intelligence in education (AIED) that can automatically track the learning status of students at each stage. KT  has been widely used in intelligent tutoring systems, adaptive learning systems and educational gaming \cite{chrysafiadi2022cognitive,hooshyar2022gamedkt}. Recently, deep learning-based methods have significantly improved the performance in KT tasks; however, this improvement comes at the cost of interpretability \cite{arrieta2020explainable,gunning2019darpa}. A lack of explainability is not conducive for stakeholders to understand the reasons behind an algorithm’s decisions, which may reduce stakeholders’ trust in these tools. For instance, if a knowledge tracing model yields unrealistic predictions, teachers may fail to understand the actual knowledge level of their students, and students may not receive an accurate assessment of their weaknesses. In addition, it is not easy for users or regulators to find defects in black box applications, which may raise security issues, such as learner resistance \cite{abdelrahman2023knowledge} or an increase in high-risk students \cite{diedrick2023adverse}. To solve the above issues, researchers have been attempting to improve the interpretability of AI in various educational tasks, such as explainable learner models \cite{long2017enhancing,castiglioni2021ai,clancey2021methods}, explainable recommender systems \cite{barria2021explainable}, explainable at-risk student prediction \cite{jang2022practical,melo2022use}, and explainable personalized interventions \cite{hur2022using}.

To the best of our knowledge, there has not been a comprehensive survey of related research on the interpretability of knowledge tracing. This paper aims to fill this gap. Compared to the existing knowledge tracing surveys \cite{2021A,liu2022knowledge,song2022survey,abdelrahman2023knowledge}, this study primarily focuses on explainable algorithms and knowledge tracing interpretability. The motivation behind this paper is threefold. First, it aims to provide a detailed and comprehensive review of the research on explainable knowledge tracing (xKT). Second, different interpretability methods for knowledge tracing should be compared, and evaluation methods should be explored. Finally, this study aims to provide a foundational and inspirational resource for researchers interested in the field of explainable knowledge tracing.

\subsection{Contributions}
The contributions of this survey are to provide an in-depth examination of the current status of explainable knowledge tracing. By doing so, it aims to establish a solid foundation of understanding and inspire additional research interest in this rapidly growing area.

Inspired by the classification criteria of xAI for complex object models as delineated by Arrieta et al. \cite{arrieta2020explainable}, this paper offers a novel categorization of knowledge tracing models into two distinct types: transparent models and black-box models. This dichotomy is further explored with a detailed examination of interpretable methods tailored to these models across three critical stages: ante hoc, post hoc, and other dimensions. 

Moreover, the current evaluation methods for explainable knowledge tracing are still lacking. In this paper, contrast and deletion experiments are conducted to explain the prediction results of the deep knowledge tracing model on the same dataset by using three XAI methods. Furthermore, this work extends an insightful overview into the evaluation of explainable knowledge tracing, tailored to varying target audiences, and delves into the prospective directions for the future development of explainable knowledge tracing. 
\begin{figure}[t]%
\centering
\includegraphics[width=0.8\textwidth]{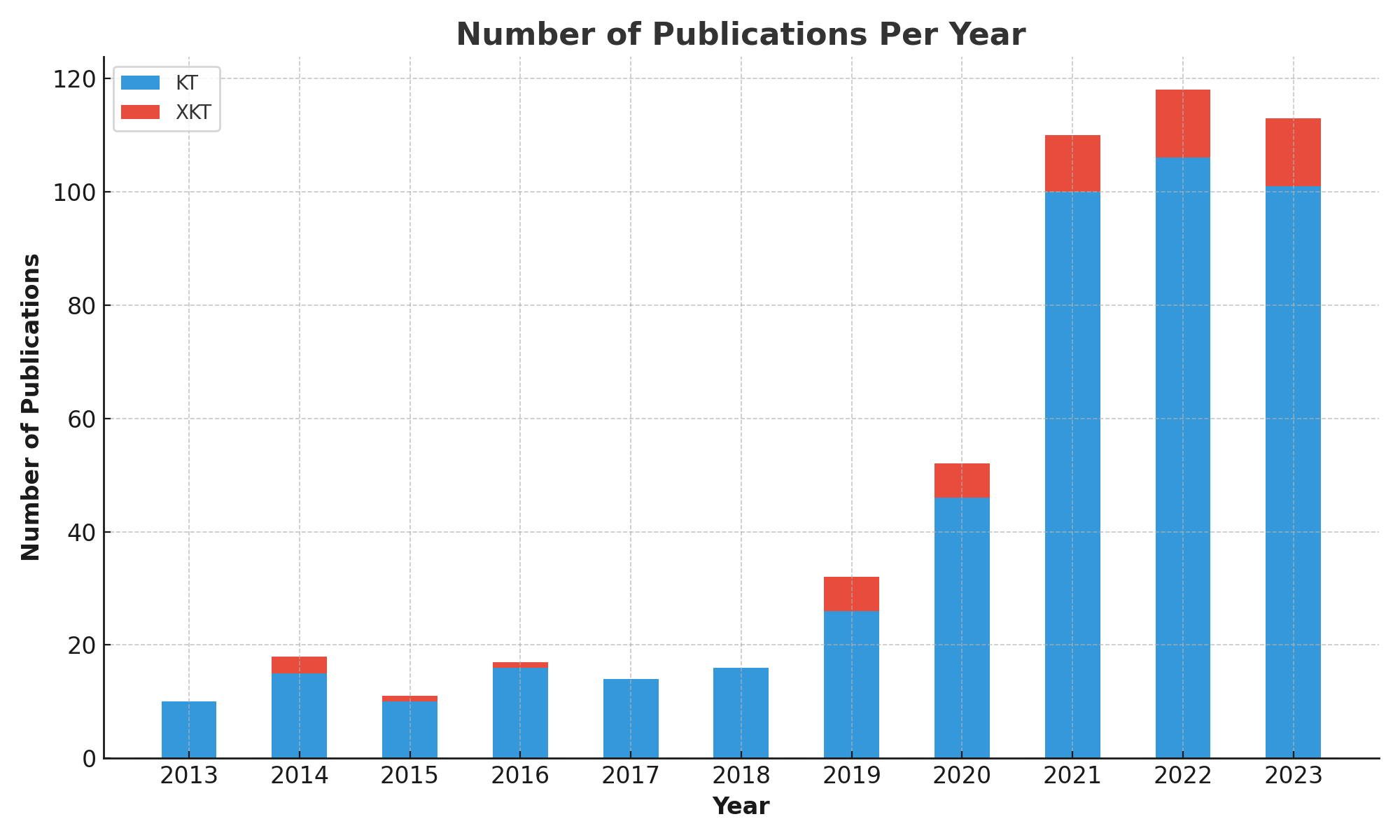}
\caption{The number of publications per year: Knowledge tracing and knowledge tracing with interpretability.}\label{pub}
\end{figure}

\subsection{Systematic Literature Review (SLR) and Execution}
In this paper, the systematic literature review (SLR) methodology was adopted. This methodology was developed by Kitchenham and Charters \cite{keele2007guidelines} and is specifically designed for comprehensive analyzes in software engineering and computer science. The SLR methodology is significant because of its systematic approach to collating and synthesizing literature, providing a comprehensive understanding of the interpretability of knowledge tracing. The survey process commenced with the formulation of specific research questions, focusing on the application and implications of xAI in knowledge tracing. The primary questions addressed were as follows: a) How can the interpretability of knowledge tracing algorithms be improved?   b) What are the applications and classifications of xAI in knowledge tracing? c) How can explainable knowledge racing models be effectively evaluated? 

Our search strategy involved a comprehensive list of keywords, such as “explainable artificial intelligence”, “xAI”, "explainable", "explainability", "explanation", "interpretable"and “knowledge tracing”. These keywords were chosen based on their prevalence in the current literature and relevance to our research questions. We used the Boolean operators ’AND’ and ’OR’ to construct detailed search strings. We conducted searches in databases such as IEEE Xplore, ACM Digital Library, Science Direct, and SpringerLink. These databases were selected for their extensive coverage of computer science and AI literature. From an initial pool of 1,783 studies, 517 were screened based on their relevance to knowledge tracing. After a full-text review, 57 papers were selected based on our inclusion and exclusion criteria, and the search period ended on November 2023.

\begin{figure}[t]%
\centering
\includegraphics[width=\textwidth]{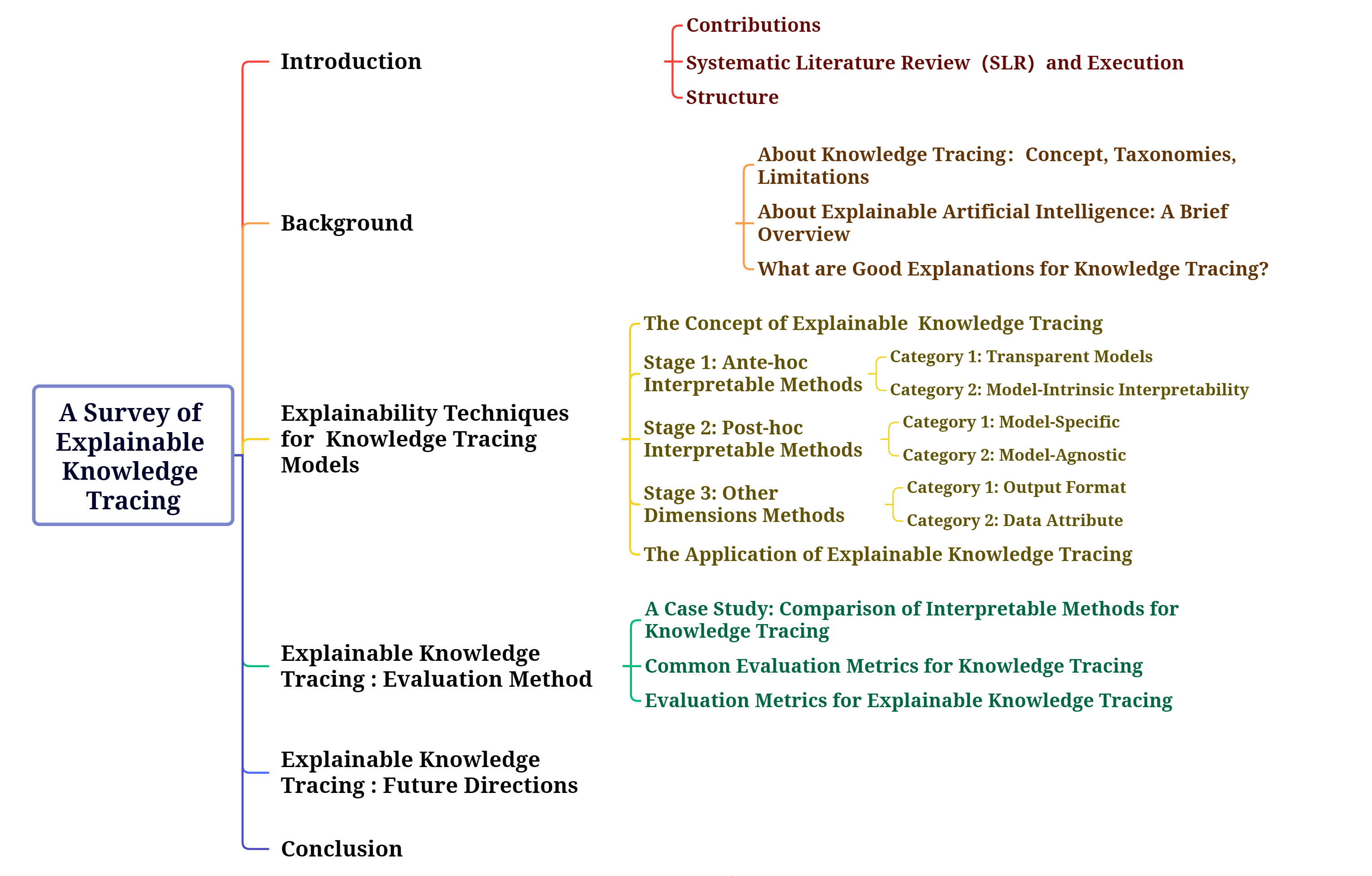}
\caption{The concept map of this survey.}\label{frame}
\end{figure}

Inclusion Criteria: 1) Relevance: Studies must focus on the interpretability of knowledge tracing algorithms, including theoretical analyses, application case studies, and empirical evaluations. 2) Novelty and Contribution: Studies should offer novel insights or approaches in the field of knowledge tracing interpretability. This includes introducing new methodologies, providing unique theoretical perspectives, or presenting novel empirical findings that significantly advance the understanding of the topic. 3) Publication Quality: Studies must be published in peer-reviewed journals or conference proceedings. In cases where no peer-reviewed version is available, but the study is highly relevant to the research topic, non-peer-reviewed versions (e.g., arXiv preprints) will also be considered for inclusion. 4) Language: The study must be written in English. \textbf{Exclusion Criteria:} 1) Relevance: Studies not directly addressing the research questions, specifically those not focusing on the interpretability of knowledge tracing algorithms, will be excluded. 2) Empirical and Methodological Rigor: Studies lacking empirical data support or detailed methodological descriptions will be excluded. This includes opinion pieces or conceptual framework studies without specific empirical analysis.

The number of papers published each year is shown in Fig. \ref{pub}. Research on the interpretability of knowledge tracing has shown a significant increasing trend since 2019, indicating that researchers have realized that high accuracy alone is insufficient to gain the trust of stakeholders when applying AI models to real-world educational scenarios, and improving the interpretability of model decisions is a crucial issue that needs to be addressed.

\subsection{Structure}
The remainder of this survey is organized as follows. Section \ref{sub2} discusses the relevant research on explainable artificial intelligence and knowledge tracing while also emphasizing the importance of interpretability in knowledge tracing algorithms. Section \ref{sub3} delves further into explainable knowledge tracing, offers insights into its various dimensions and implications, and presents a detailed examination of interpretable methods suitable for explaining knowledge tracing models. The focus of Section \ref{sub4} is on the methodologies for scientifically evaluating the interpretability of knowledge tracing models. Finally, Section \ref{sub5} outlines future research directions in explainable knowledge tracing, pinpointing key areas that warrant further investigation and innovation. To enhance readers’ understanding of this paper’s architecture, we depict it in detail in Fig. \ref{frame}.

\section{Background}\label{sub2}
This section thoroughly examines the developmental background of KT and xAI, presenting the latest frameworks and methodologies in these fields. It delves into the concepts, classifications, and evolution of KT models while revealing the inherent limitations of these models. The section also explores the ongoing challenge of finding a balance between model accuracy and interpretability, discussing how to achieve the optimal compromise between the two. Additionally, it offers an in-depth analysis of xAI's fundamental principles, methodologies, and evaluation mechanisms, underscoring its crucial role in enhancing transparency and comprehension of complex AI systems. Specifically, the section explores the application of tailored explanation methods in the KT domain, proposing targeted solutions for both professional and lay users. By comprehensively analyzing the theoretical foundations and practical applications of KT and xAI, this chapter aims to provide groundbreaking insights into the interpretability of knowledge tracing.

\subsection{About Knowledge Tracing: Concepts, Taxonomies, Evolution, Limitations}
Knowledge tracing has become a key component of learner models. A large amount of historical learning trajectory information provided by an intelligent tutoring system (ITS) is used to model learners’ knowledge states and predict their performance in future exercises \cite{qiu2023okpt}. Thus, knowledge tracing provides personalized learning strategies \cite {abdelrahman2023learning} and learning path recommendations \cite{ma2023personalized} for education stakeholders and is a crucial element of adaptive education.
Specifically, knowledge tracing is a task for predicting students’ performance in future practice according to changes in learners’ knowledge mastery in historical practice \cite{2015Deep}; this task involves two main steps: 1) modeling learners’ knowledge state according to their historical practice sequence and 2) predicting learners’ performance in future practice. In other words, this task can be formulated as a supervised time series learning task, where $X_i = \{e_t,r_t\}$ represents a student’s answer pair, $e_t$ represents the exercise ID, and $r_t$ represents the answer result for related exercise $e_t$, $r_t$ $\in$ $\{0, 1\}$ (1 indicates the correct answer and vice versa). Given a student’s exercise sequence $X = \{x_1, x_2, x_3,\ldots, x_{(t-1)}\}$ and the next exercise, the task objective is to predict the correct probability $P(r_t = 1 | X, e_t )$ of the exercise $e_t$.

\begin{table*}[t]
    \caption{The taxonomies of knowledge tracing.}
    \label{KT}
    \resizebox{\textwidth}{!}{
    \footnotesize
    \begin{tabular}{m{3cm}m{7cm}m{4cm}}
    \toprule
      Category         & Description              & Representative Works                    \\
    \midrule
     Markov process-based knowledge tracing      & Assuming that a student's learning process is representable as a Markov process, it can be modeled with probabilistic models.   & BKT \cite{Albert1994Knowledge} DBKT \cite{kaser2017dynamic}      \\  \hline
     Logistic knowledge tracing                  & The logistic function represents the probability of a student answering an exercise correctly, under the premise that this probability is expressible as a mathematical function involving the student and the KC parameter. The logistic model posits that students' binary answers (correct or incorrect) adhere to Bernoulli distributions.  & LFA \cite{cen2006learning} PFA \cite{pavlik2009performance} KTMs \cite{vie2019knowledge} \\ \hline
     Deep learning-based knowledge tracing       & DLKT models can simulate changes in students' knowledge states and encapsulate a broad spectrum of complex features, which might be challenging to extract through other methods. & DLKT: \cite{2015Deep} \cite{zhang2017dynamic} \cite{su2018exercise} \cite{DBLP:conf/edm/PandeyK19} \cite{ghosh2020context} \cite{nakagawa2019graph} \cite{pandey2020rkt} \cite{liu2019ekt} \cite{minn2022interpretable} \\
    \bottomrule
    \end{tabular}}
\end{table*}

\begin{figure*}[t]
    \centering
    \includegraphics[width=\textwidth]{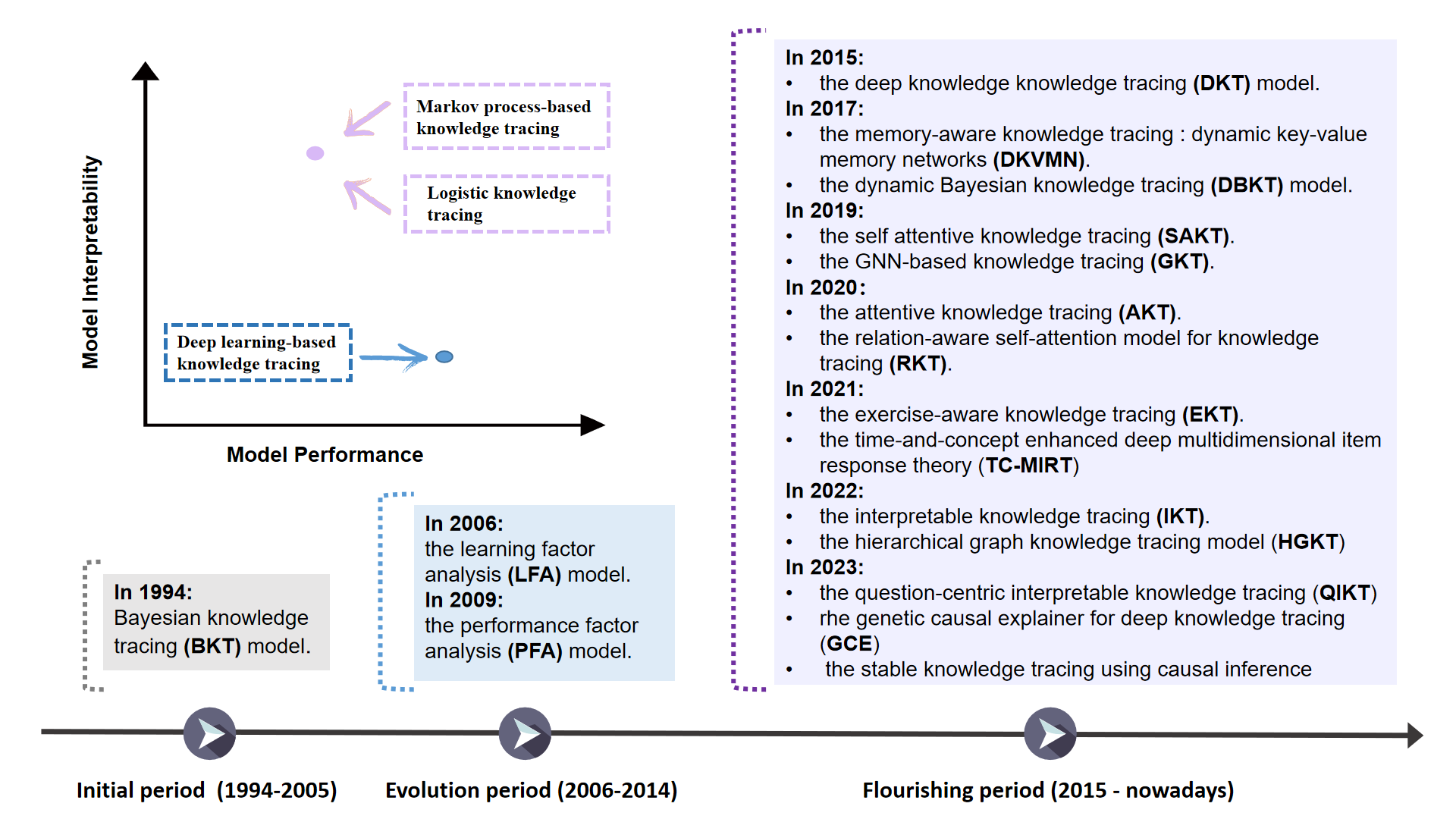}
    \caption{\textrm{The timeline of seminal works toward knowledge tracing.}}
    \label{eva}
\end{figure*}

According to the general classification method of knowledge tracing models, existing models can be classified into the following three categories \cite{liu2022knowledge}: 1) Markov process-based knowledge tracing, 2) logistic knowledge tracing, and 3) deep learning-based knowledge tracing (DLKT). The taxonomies of knowledge tracing are shown in Table \ref{KT}. Next, we introduce a series of seminal works on the aforementioned three types of models and outline the timeline of knowledge tracing evolution, as shown in Fig. \ref{eva}.

In 1994, Corbett et al. proposed Bayesian knowledge tracing (BKT) \cite{Albert1994Knowledge}, which is based on a two-state Hidden Markov Model(HMM) that treats student knowledge states as hidden variables \cite{1966Statistical}. However, since the model uses a shared set of parameters for the same knowledge component (KC), it cannot personalize modeling for students at different levels. To overcome this limitation, researchers have added personalized features to make the model more realistic, leading to the emergence of various variations based on Bayesian knowledge tracing, marking the initial phase of knowledge tracing research. One notable improvement is dynamic BKT (DBKT) \cite{kaser2017dynamic}. To address the issue of BKT modeling each KC individually, DBKT employs dynamic Bayesian networks to represent multiple KCs jointly in a single model. This approach models the prerequisite hierarchies and relationships within KCs. Both DKT and DBKT are representative models of knowledge tracing based on the Markov process.
In 2006, Cen et al. \cite{cen2006learning} proposed learning factor analysis (LFA), which inherits the Q matrix used in psychometrics to assess cognition and extends the theory of learning curve analysis. An improved LFA model is performance factor analysis (PFA) \cite{pavlik2009performance}, which was developed in 2009. Additionally, knowledge tracing machines (KTMs) \cite{vie2019knowledge} utilize a factorization machine to model all variable interactions. These three methods are representative models of logistic knowledge tracing, and a detailed explanation of each will be provided in Section \ref{sub3.2}. In general, logistic knowledge tracing has achieved better performance than BKT, and knowledge tracing has gradually entered a development period \cite{Schmucker_Wang_Hu_Mitchell_2022}.
Since deep knowledge tracing (DKT) \cite{2015Deep} was proposed in 2015, deep learning techniques have shown more vital feature extraction ability in knowledge tracing than the other two types of models. Based on this seminal work, much deep learning-based knowledge tracing (DLKT) has emerged. For example, researchers have applied deep learning techniques to knowledge tracing in various ways. Below, we list several categories of representative work:  1) memory-aware knowledge tracing: DKVMN \cite{zhang2017dynamic};  2) attention-aware knowledge tracing: SAKT \cite{DBLP:conf/edm/PandeyK19} and AKT \cite{ghosh2020context}; 3) graph-based knowledge tracing: GKT \cite{nakagawa2019graph} and HGKT \cite{tong2022introducing}; 4) relation-aware knowledge tracing: RKT \cite{pandey2020rkt}; 5) exercise-aware knowledge tracing: EKT \cite{liu2019ekt}; and 6) interpretable knowledge tracing: TC-MIRT \cite{su2021time}, IKT \cite{minn2022interpretable}, QIKT \cite{DBLP:conf/aaai/00060HL023}, GCE \cite{li2023genetic}, and stable knowledge tracing using causal inference \cite{zhu2023stable}.

As shown in Fig. \ref{eva}, before the emergence of deep learning in knowledge tracing, Bayesian knowledge tracing and logistic knowledge tracing were widely used because of their relatively simple model structures and powerful interpretability \cite{10.1145/3375627.3375856}. However, due to the massive and multidimensional nature of online learning data, these two types of models were unable to achieve good performance on big data \cite{10.3102/0091732X20903304}. Deep learning-based models have a clear advantage in processing large datasets. However, when applied to real-world teaching scenarios, DLKT may face the following challenge: the large number of network layers and parameters in deep networks may limit the interpretability of the generated parameters. Additionally, the lack of interpretability in deep learning-based models can also lead to potential ethical and privacy issues. Stakeholders need to be able to trust the models and understand how the models make decisions.

Overall, DLKT models exhibit strong performance but poor interpretability {\cite{abdelrahman2023knowledge}, while simple models with strong interpretability are far weaker than the former, as shown in the upper left corner of Fig. \ref{eva}. Consequently, the tradeoff between interpretability and performance poses a significant challenge for researchers challenge for researchers. In recent years, researchers have utilized various methods to explain knowledge tracing models and have attempted to maximize transparency while ensuring model performance. In Section \ref{sub3}, we will elaborate on the interpretable methods existing in the above proposed models.


\begin{figure*}[t]
    \centering
    \includegraphics[width=\textwidth]{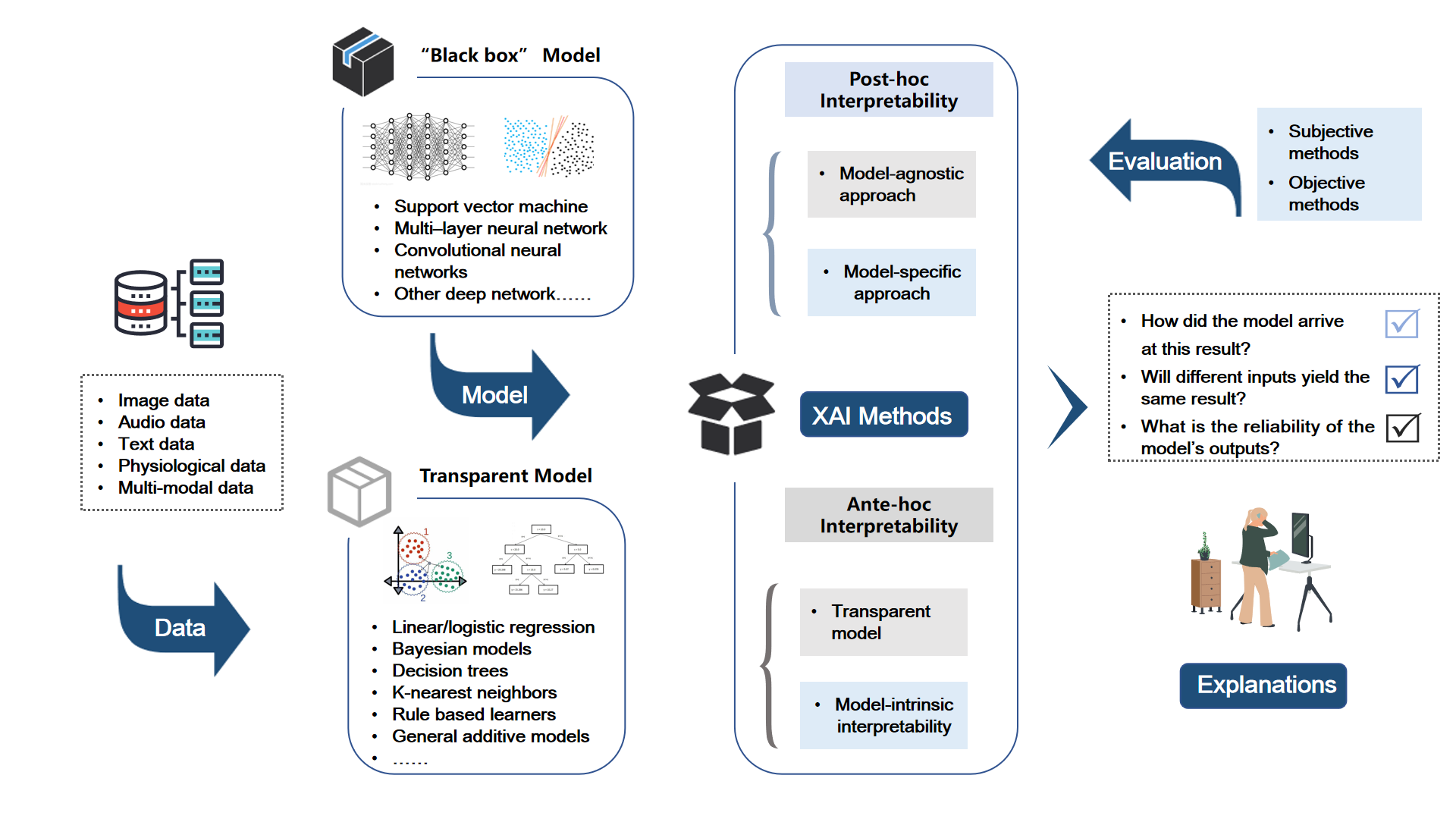}
    \caption{\textrm{The framework of explainable artificial intelligence (xAI).}}
    \label{xAI}
\end{figure*}

\subsection{About Explainable Artificial Intelligence (xAI): A Brief Overview} \label{sub2.2}
The goal of explainable artificial intelligence (xAI) is to provide an understanding of the internal workings of a system in a manner that humans can comprehend. XAI aims to answer questions such as “How did the model arrive at this result?”, “Will different inputs yield the same result?”, and “What is the reliability of the model’s outputs?”. In essence, xAI’s purpose is to provide an explanation to the explainees regarding why the model generates the corresponding output based on the input. Based on the diverse needs of explainees, explainers offer appropriate types of explanations. The process of explanation provided by xAI enhances explainees’ degree of trust in the system, thus increasing the system’s utility ratio across various industries. To enhance the readers’ understanding of explainable artificial intelligence, we introduce the xAI framework illustrated in Fig. \ref{xAI}. Improving the interpretability of algorithms is important in AIED, and the benefits can be summarized as follows: 1) Developers can enhance the transparency of models in a more scientific way, which can lead to better model optimization. 2) Transparency can help domain experts discover the cognitive rules in the learning process, leading to deeper insights and better decision-making. 3) Transparency can help users better understand the reasons and logic behind AI-driven decisions, which can increase their trust in the technology. 4) Regulatory authorities can use transparency to achieve effective supervision and ensure the safety of intelligent products used in education while also ensuring compliance with the law.

\begin{table*}[t]
    \caption{A summary of representative interpretable methods.}
    \label{PP}
    \centering
    \resizebox{\textwidth}{!}{
    \small
    \begin{tblr}{
        colspec={Q[l,m,0.25\textwidth] Q[c,m,0.1\textwidth] Q[c,m,0.1\textwidth] Q[c,m,0.08\textwidth] Q[c,m,0.08\textwidth] Q[c,m,0.08\textwidth] Q[c,m,0.1\textwidth] Q[l,m,0.7\textwidth]},
        hspan=minimal,
        rowsep=0.6pt,
        vline{2,3}={1-Z}{0.5pt},vline{5}={1-Z}{0.5pt},vline{7}={1-Z}{0.5pt},vline{8}={1-Z}{0.5pt}
       }
    \hline[1pt]
    \SetCell[r=2]{l,m}{Methods}  & \SetCell[r=2]{c,m}{Year}  & \SetCell[c=2]{c,m}{Stage} &   & \SetCell[c=2]{c,m}{Category} &  &\SetCell[r=2]{c,m}{Domain}  & \SetCell[r=2]{l,m}{Description}\\\cline{3-6}  
           & &Anc-hoc &Post-hoc &Model-specfic & Model-Agnostic & &                        \\                                  
    \hline
    Attention\cite{DBLP:journals/corr/BahdanauCB14}    &   2014   & \checkmark &    & \checkmark &  & CV/NLP  & Attention weight matrix visualization\\
    Bayes Rule List\cite{letham2015interpretable}    &   2015   & \checkmark &    & \checkmark &  & -  & Trees and Rule-based Models\\
    Generalized additive models (GAMs)\cite{caruana2015intelligible}    &   2015   & \checkmark &    & \checkmark &  & -  & The final decision form is obtained by combining each single feature model with linear function\\
    Neural Additive Model\cite{agarwal2021neural}   &   2020   & \checkmark &    & \checkmark &  & CV  & Train multiple deep neural networks in an additive fashion such that each neural network attend to a single input feature\\
    Activation Maximization\cite{erhan2010understanding}    &   2010   &  &\checkmark    & \checkmark &  & CV  & Maximize neuronal activation by identifying the optimal input for a neuron at a specific network layer\\
    Gradient-based Saliency Maps\cite{DBLP:journals/corr/SimonyanVZ13}    &   2013   &  &\checkmark    &  &\checkmark   & CV  &\SetCell[r=2]{l,m}{The back propagation mechanism of DNN is used to propagate the decision importance signal of the model from the output layer neurons to the input of the model layer by layer to deduce the feature importance of the input samples} \\
    DeConvolution Nets\cite{zeiler2014visualizing}    &   2014   &  &\checkmark    &  &\checkmark & CV  & \\
    Guided Backprops\cite{DBLP:journals/corr/SpringenbergDBR14}    &   2015   &  &\checkmark    &  &\checkmark  & CV  & \\
    SmoothGrad\cite{smilkov2017smoothgrad}    &   2017   &  &\checkmark    &  & \checkmark & CV  & \\
    Layer-wise Relevance BackPropagation (LRP)\cite{bach2015pixel}    &   2015   &  &\checkmark    &  &\checkmark  & CV/NLP  & \\
    Salient Relevance (SR) Map\cite{li2019beyond}    &   2019   &  &\checkmark    &  &\checkmark  & CV  & \\
    Class Activation Mapping (CAM)\cite{zhou2016learning}    &   2016   &  &\checkmark    &  &\checkmark  & CV  &\SetCell[r=2]{l,m}{The neural network's feature map is utilized to ascertain the significance of each segment of the original image}  \\
    Grad-CAM\cite{selvaraju2017grad}    &   2017   &  &\checkmark    &  &\checkmark  & CV  &  \\
    Grad-CAM++\cite{chattopadhay2018grad}    &   2018   &  &\checkmark    &  &\checkmark  & CV  &  \\
    Local Interpretable Model-Agnostic Explanations (LIME)\cite{ribeiro2016should}    &   2016   &  &\checkmark    &  &\checkmark  & CV  &An interpretable model with simple structure is used to locally approximate the decision result of the model to be explained for an input instance \\
    SHapley Additive exPlanations (SHAP)\cite{lundberg2017unified}    &   2017   &  &\checkmark    &  &\checkmark  & CV  & Reflects the influence of each feature in the input sample and shows the positive and negative influence  \\
    Concept Activation Vectors(CAV)\cite{kim2018interpretability}    &  2018   &  &\checkmark    &  &\checkmark  & CV  & Measures the relatedness of concepts within the model's output  \\                                \hline[1pt]
    \end{tblr}}
\end{table*}

Existing algorithms can be classified as transparent (white box) or black box models, depending on their complexity \cite{guidotti2018survey} (details in Section \ref{sub3.1}). Transparent models are characterized by simple internal components and self-interpretability, allowing users to intuitively understand their internal operation mechanism \cite{NEURIPS2022_65398a0e}. A black box model refers to a model with a complex, nonlinear relationship between the input and output, with an operating mechanism that is difficult to understand, such as that of a neural network \cite{Zhang9472817}. Based on the characteristics of these two models, xAI methods are typically categorized as ante-hoc interpretable methods or post-hoc interpretable methods \cite{kamath2021explainable}.
Ante-hoc interpretable methods are mainly applied to models with simple structures and strong interpretability (such as transparent models) or to build interpretable modules in the model to make it intrinsically interpretable. In contrast, post-hoc interpretable methods, such as black box models, develop interpretive techniques to interpret trained machine learning models. Post-hoc interpretable methods are usually subdivided into model-specific approaches and model-agnostic approaches based on their application scope \cite{2018Peeking}. These methods are introduced in detail in the following sections. Table \ref{PP} provides an overview of representative interpretable methods.

There is currently no widely accepted scientific evaluation standard for xAI. Different experts from various disciplines have conducted preliminary investigations based on different evaluation objectives, such as the characteristics of the model being evaluated or the requirements of users and application scenarios \cite{mohseni2021multidisciplinary}. One prominent evaluation method is the three-level approach proposed by Doshi Velez et al. \cite{doshi2018considerations}, which includes the following steps: 1) application-grounded evaluation, 2) human-grounded evaluation, and 3) functionally-grounded evaluation. These three levels of evaluation provide useful frameworks for evaluating the interpretability of xAI systems in different contexts.

\begin{table*}[t]
    \caption{Classification of xAI Evaluation Methods Based on User Involvement}.
    \label{ext}
    \resizebox{\textwidth}{!}{
    \footnotesize
    \begin{tblr}{
        colspec={Q[l,m,0.1\textwidth] Q[l,m,0.4\textwidth] Q[l,m,0.4\textwidth]},
        hspan=minimal,
    }
    \hline[1pt]
      Angles         & Methods              & Limitations                    \\
    \hline
     Subjective evaluation     & Qualitative evaluation based on open-ended questions\cite{hoffman2018metrics,markus2021role,balog2020measuring}; \newline Quantitative evaluation based on closed-ended questions\cite{lage2019human}, \cite{DBLP:journals/jbi/MarkusKR21};\newline A mixed-methods approach that combines both qualitative and quantitative evaluation.\cite{kumarakulasinghe2020evaluating,vilone2021notions} & This method may be susceptible to bias and variability owing to individual differences among evaluators. Moreover, the requisite for professional human resources can lead to elevated evaluation costs. \\ 
      Objective evaluation     & Fidelity \cite{alvarez2018towards}; Consistency\cite{robnik2018perturbation}; Stability\cite{vilone2021notions};  Sensitivity\cite{vilone2021notions};  Causality\cite{moraffah2020causal}; Complexity \cite{wu2018beyond,fan2020interpreting} & Models might rely on particular evaluation methods, and varying metrics could yield disparate results.  \\
    \hline[1pt]
    \end{tblr}
    }
\end{table*}
Generally, application-level evaluation is considered an effective approach because the interpretation is applied to the appropriate field and evaluated by professionals, resulting in more convincing results. Currently, the popular classification method for xAI evaluation involves dividing it into subjective and objective evaluations based on whether humans are involved\cite{vilone2021notions,coroama2022evaluation}. Table \ref{ext} shows the evaluation methods, metrics, and current limitations of both subjective and objective evaluation in xAI. It is important to carefully select appropriate evaluation metrics based on the specific features and goals of the evaluated xAI system. Overall, the combination of subjective and objective evaluation methods may be the most effective approach for assessing the interpretability of xAI systems while balancing cost and performance.

\subsection{What are good explanations for knowledge tracing?} \label{sub2.3}
According to Merriam-Webster, ”interpret” means to present something in understandable terms and explain its meaning \cite{merriam-webster}. However, what constitutes a good explanation varies across different fields, and experts have attempted to define it in different ways. In computer science, Lipton \cite{1990Contrastive} emphasized the importance of comparative explanations, i.e., whether the predicted outcome $Y$ will change for different inputs $X$. Physicist Max Tegmark described a good explanation as one that answers more questions than asked \cite{brockman2013explains}. Moreover, psychology researchers have highlighted the significance of explanations in learning and inference and how individuals’ explanatory preferences can impact explanation-based processes in a systematic way \cite{Lombrozo2016Explanatory}. In certain scenarios of adaptive education, researchers have used verbal and visual explanations \cite{williamson2021effects} or interactive interfaces \cite{ghai2021explainable} to provide explanations and have achieved positive outcomes. For example, Cristina Conati et al. \cite{conati2021toward} added an interactive simulation program to the adaptive CSP (ACSP) applet to provide an explanation function. The research results demonstrated that providing explanations can enhance students’ trust in ACPS prompts. 

Explanations also play a pivotal role in the process of knowledge tracing, but what are good explanations for knowledge tracing?  Research suggests that explanations must be audience-specific and goal-oriented \cite{phillips2020four,mohseni2021multidisciplinary}.Stakeholders in knowledge tracing are divided into professional users (developers, researchers) and non-professional users (teachers, students), each requiring tailored explanations \cite{ferreira2020people}. For professionals, explainability enhances understanding, system debugging, model optimization, and credibility 
\cite{khosravi2022explainable}. For non-professionals, it facilitates comprehension of learning processes, encourages result acceptance, and boosts model satisfaction \cite{hoffman2023measures}. The field has developed methods ensuring both accuracy and transparency, making model operations and decisions clear to all users, thereby improving interpretability. Further details on these methods will be provided in the next section.

\section{Explainability Techniques for Explainable Knowledge Tracing Models}\label{sub3}
This section delves into the key aspects of enhancing the interpretability of knowledge tracing models. Initially, we categorize and discuss explainable knowledge tracing models, focusing on the critical distinctions between transparent models and complex black-box models. This discussion lays the groundwork for understanding the internal mechanisms of these models. Subsequently, we shift our focus to exploring methods for augmenting the interpretability of knowledge tracing. These methods encompass both ante-hoc and post-hoc strategies, as well as other dimensions. The aim is to reveal how various approaches can enhance the models' transparency and comprehensibility. Finally, we examine the practical implementation of explainable knowledge tracing in real-world applications, such as generating diagnostic reports. The section concludes with a critical discussion evaluating the balance between the models' interpretability, accuracy, and their practical application in educational settings.

\subsection{The Concept of Explainable Knowledge tracing}\label{sub3.1}
As mentioned in Section \ref{sub2.2}, machine learning models are typically categorized as transparent or black box models according to the complexity of the objects they are intended to explain, based on the criteria of xAI. Transparent models are characterized by high transparency of internal components and self-interpretability, such as, linear/logistic regression \cite{harrell2001regression,2002An}, bayesian models \cite{carlin1995bayesian,lye2021sampling,raftery1995bayesian}, decision trees \cite{1987Simplifying,1976Constructing}, k-nearest neighbors \cite{guo2004knn,nefeslioglu2010assessment,imandoust2013application}, rule based learners \cite{setnes1998rule,nunez2002rule,nunez2006rule}, general additive models \cite{hastie1987generalized,hastie2017generalized,wood2006generalized}, etc. For transparent models, interpretability can be understood from three perspectives: algorithmic transparency, decomposability, and simulatability \cite{lipton2018mythos}. However, models such as multi–layer neural network \cite{2016DeepRED,2016Not} and other deep network \cite{2015Show,2016DeepRED,2020Grad}, which have complex internal structures and difficult operation mechanisms, are usually referred to as black box models. XAI algorithms are capable of understanding the architecture and layer configurations of transparent models, but they lack the ability to comprehend the operational mechanisms inherent in black box models\cite{das2020opportunities}.



\begin{figure*}[t]
    \centering
    \includegraphics[width=\textwidth]{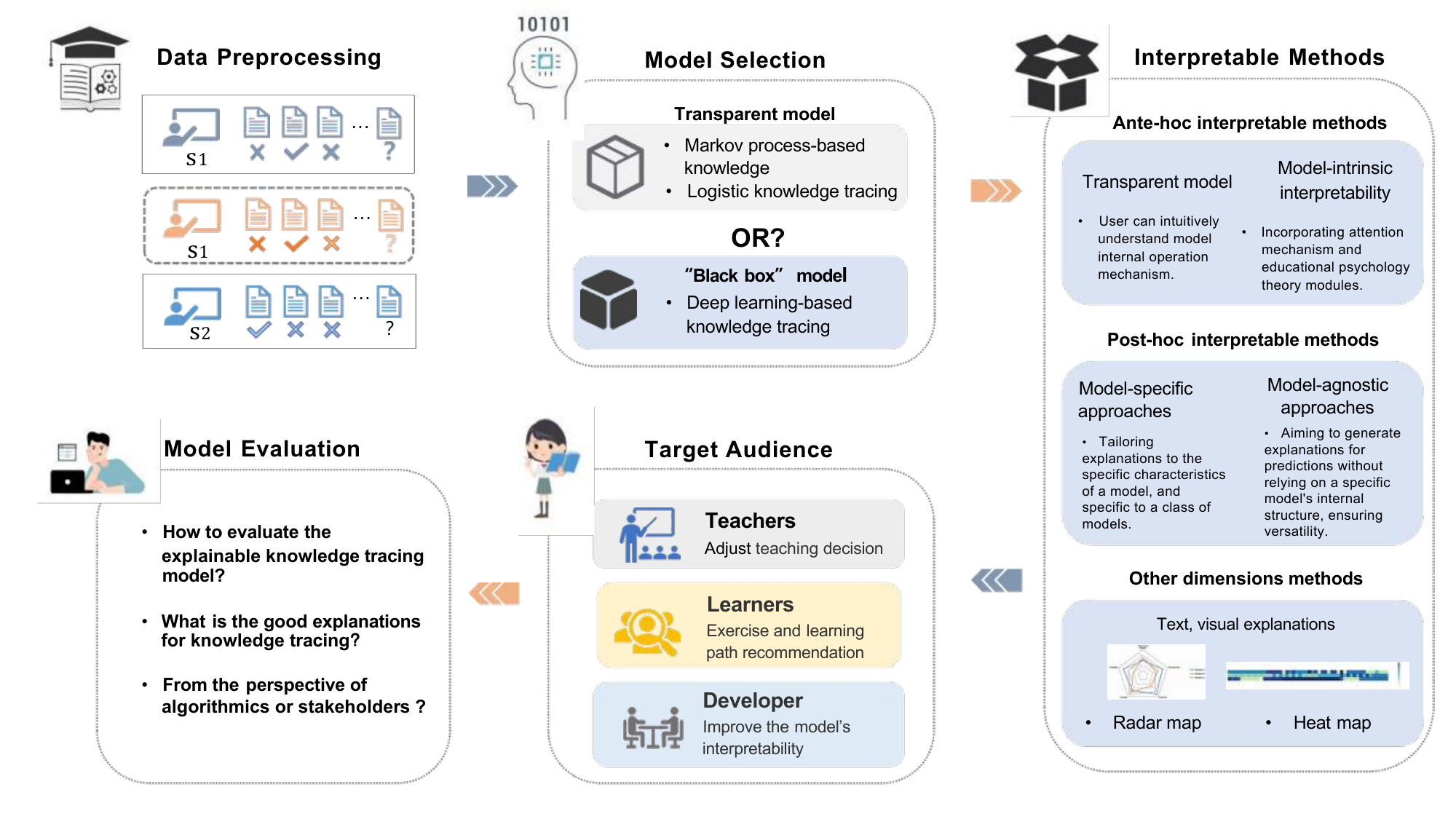}
    \caption{\textrm{The framework of explainable knowledge tracing(xKT).}}
    \label{xKT}
\end{figure*}

\textbf{Explainable Knowledge Tracing Model Taxonomy.} 
Inspired by the criteria for classifying model complexity in xAI \cite{arrieta2020explainable}, we propose a novel taxonomy specifically tailored for knowledge tracing models. This framework categorizes models based on their explainability and comprehensibility. We identify models employing Markov processes and logistic regression as "transparent models" due to their straightforward structures and the ease with which users can understand them. This classification is rooted in the models' interpretability and the transparency of their decision-making processes, emphasizing the accessibility and interpretability of how decisions are made. For example, state transitions in Markov models and parameter settings in logistic regression models are intuitive, making these models' decision-making processes easily traceable and explainable. In contrast, knowledge tracing models based on deep learning, especially those involving complex multi-layered network structures, are considered akin to "black box" by users. The internal mechanisms of these models are difficult to comprehend because of their complexity and rich nonlinear characteristics, obscuring the internal decision-making process. Consequently, these models and their variants are categorized as "black-box models". We define explainable knowledge tracing and show its framework in Fig. \ref{xKT}. 

\begin{figure*}[t]
    \centering
    \includegraphics[width=\textwidth]{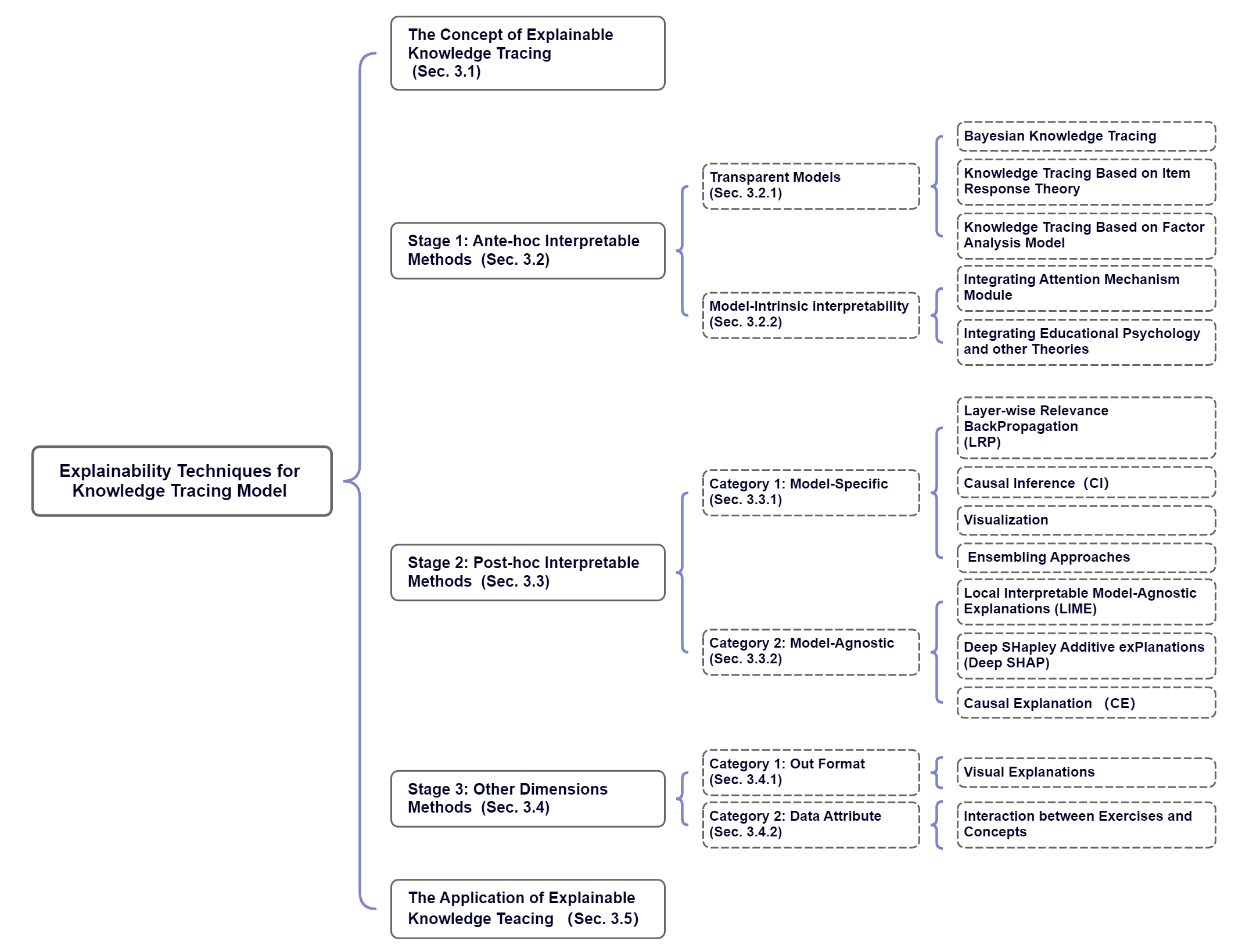}
    \caption{\textrm{The framework of explainability techniques for the knowledge tracing model.}}
    \label{tfkt}
\end{figure*}

\textbf{Methodologies for Explainable Knowledge Tracing.}
In the following section, we delve into interpretable methods for the two types of knowledge tracing models mentioned earlier, categorizing them into three phases: 1) Ante-hoc interpretable methods; 2) Post-hoc interpretable methods; and 3) Other dimensions. Ante-hoc interpretable methods that focus on transparent models and model-intrinsic interpretability, which can be achieved by simplifying the model’s structure or incorporating intuitive modules to enhance its interpretability. Post-hoc interpretability methods, on the other hand, center around model-specific and model-agnostic approaches, such as using external tools or techniques to elucidate the model’s decision-making process. Other dimensions include interpretability methods that are specific to knowledge tracing models but have not yet been widely discussed in the current xAI literature. An example would be leveraging the associative relationships between problems, concepts, or users within the context of knowledge tracing.
To aid readers in locating various interpretable methods, we present a framework diagram in Section \ref{sub3} in Fig. \ref{tfkt}. Additionally, all the reviewed explainable knowledge tracing methods are summarized in Table \ref{summary-table}. 

\subsection{Stage 1: Ante-hoc Interpretable Methods}\label{sub3.2}
As mentioned in Section \ref{sub2.2}, ante-hoc interpretable methods are primarily used for transparent models or model-intrinsic interpretability \cite{sarkar2022framework}. These methods aim to make the model itself capable of interpretation by training a transparent model with a simple structure and strong interpretability or by building interpretable components into a complex model structure \cite{arrieta2020explainable,das2020opportunities}. This paper mainly focuses on two types of ante-hoc interpretable methods for xKT: transparent models and model-intrinsic interpretability; these methods are described in the following subsections.

\subsubsection{Category 1: Transparent Models}  
Based on the classification criteria in Section \ref{sub3.1}, knowledge tracing models that use the Markov process and logistic regression are considered knowledge tracing transparent models due to their easy-to-understand internal structure and operational process. The following sections elaborate on how the knowledge tracing transparent model achieves interpretability.

\begin{figure*}[t]
    \centering
    \includegraphics[width=0.8\textwidth]{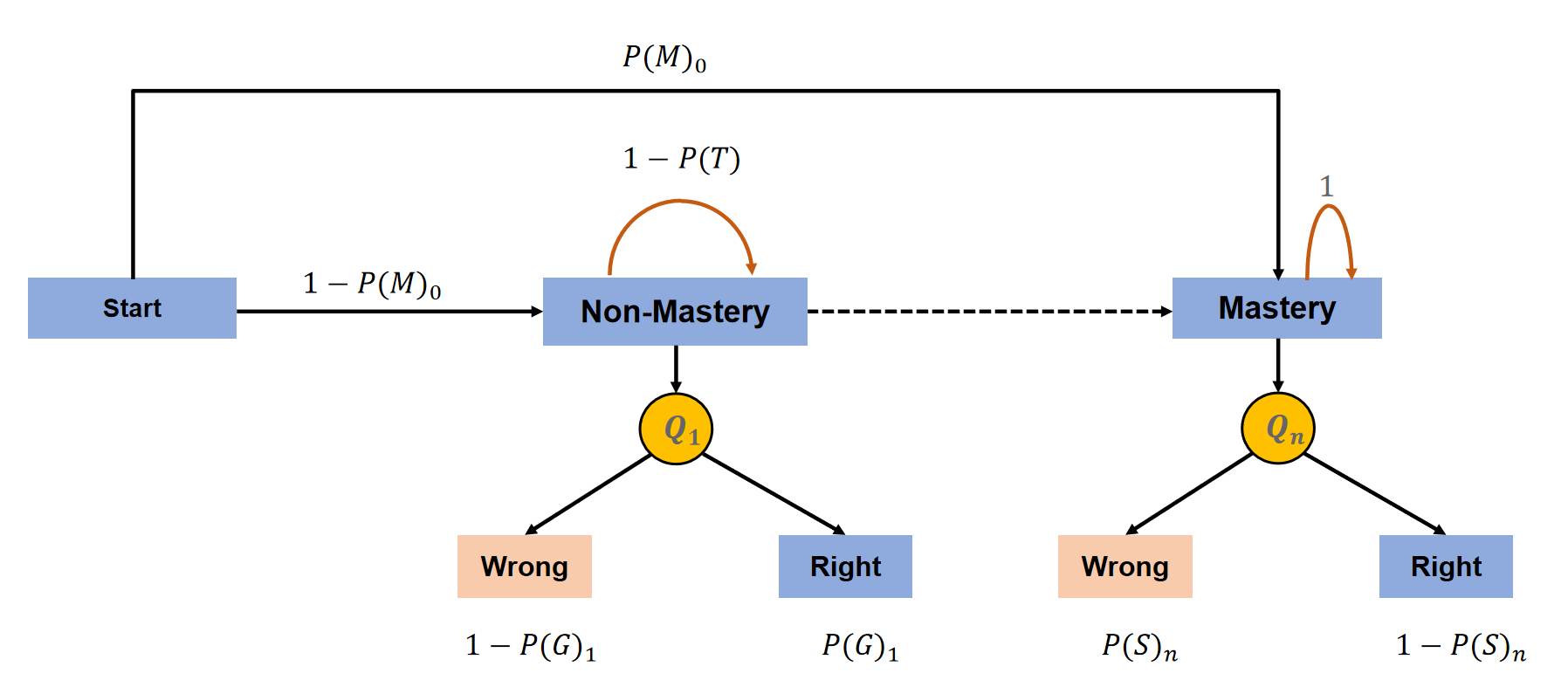}
    \caption{\textrm{State transitions for Bayesian knowledge tracing.}}
    \label{fig:3}
\end{figure*}

\textbf{Bayesian Knowledge Tracing (BKT).} BKT is a probabilistic model grounded in the Markov process, predicting students' skill mastery by updating beliefs based on performance. Although not as performant as deep knowledge tracing models, BKT's use of the HMM framework ensures a satisfactory and comprehensible explanation of the knowledge tracing. Some extended models of BKT are personalized by incorporating individual student characteristics. For example, Pardos et al. \cite{pardos2010modeling} set different initial probabilities for different students to achieve partial personalization. However, differentiating the initial probabilities only partially achieves personalization, and Lee et al. \cite{lee2012impact} used a separate set of personalization parameters for each student to adequately model interindividual differences. Although the personalized parameters are good for conferring variability across individuals, they do not consider the influence of knowledge concepts. Hawkins et al. \cite{hawkins2014using} proposed BKT-ST to calculate the similarity between current knowledge concepts and those learned previously, enhancing the model’s capacity to represent interconnected linkages across concepts. In addition, Wang et al. \cite{wang2016structured} proposed the use of multigrained-BKT and historical-BKT to model the relationships between different knowledge components (KCs). Moreover, Sun et al. \cite{sun2022genetic} used a genetic algorithm (GA) to optimize the model to solve the exponential explosion problem when tracing multiple concepts simultaneously. Moreover, Moreover, the interpretable knowledge tracing (IKT) model proposed by Minn et al. \cite{minn2022interpretable}, which is distinct in its use of tree-augmented naive Bayes and focuses on skill mastery, learning transfer, and problem difficulty, offers greater interpretability and adaptability in student performance prediction. Beyond basic parameters and knowledge concepts, the BKT model is also increasingly taking into account students' emotions and additional behavioral aspects. For examples, Spaulding et al. \cite{spaulding2015affect} introduced the concept of affective BKT, integrating students' emotional states into the model. Furthermore, to accurately capture how students' memory retention changes over time, Nedungadi et al. \cite{nedungadi2014predicting} developed the PC-BKT model. This adaptation incorporates a temporal decay function to model the process of forgetting, offering a more nuanced understanding of students' learning and memory retention behaviors. The HMM describes the probabilistic relationship between observable and hidden variables, and the probabilistic relationship varies over time. In Bayesian knowledge tracing, the model estimates students’ learning state by observing the results of students’ responses to questions related to knowledge concepts. The state transfer probability calculation process and the decision process are transparent, and the change process of the model can be effectively observed through the state transfer diagram, as shown in Fig. \ref{fig:3}. Here, \( P(M) \) represents the probability of mastery, indicating a student’s current understanding of the skill, \( P(T) \) represents the probability of transition, which is the rate at which a student transitions from nonmastery to mastery, \( P(G) \) is the probability of guessing correctly, accounting for lucky guesses, and \( P(S) \) refers to the probability of slipping, where a mastered skill is incorrectly applied. In BKT, knowledge mastery is updated along with the learning parameters. When the probability of students mastering the relevant knowledge concept in the initial knowledge state is greater than 0.95 \cite{corbett1994knowledge}, students have mastered the knowledge concept. 

\begin{figure*}[t]
    \centering
    \includegraphics[width=0.7\textwidth]{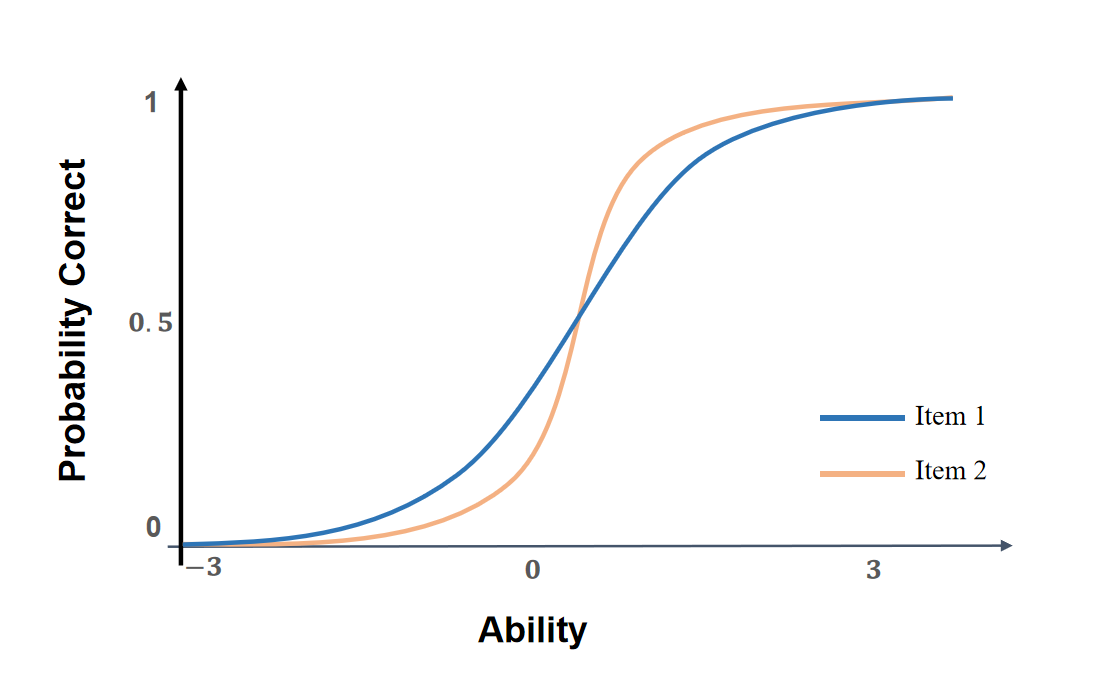}
    \caption{\textrm{Parameters in item response theory.}}
    \label{fig:3-1-1-irt}
\end{figure*}

In summary, from an interpretable perspective, the Bayesian knowledge tracing model has good transparency in the computational process as a purely probabilistic model. Observing different state transfers of BKT can provide a basis for modeling decisions for students and teachers. From the model perspective, on the one hand, Bayesian knowledge tracing models do not account for the differences in the initial knowledge levels of different students and lack an assessment of the difficulty of the questions. Although many models have been extended based on BKT, they still cannot be applied to knowledge tracing scenarios with large-scale data. On the other hand, these models assume that students do not forget the knowledge they have mastered, which is not consistent with actual cognitive characteristics. In addition, using binary groups to represent recent knowledge states does not match the real cognitive state situation. It is difficult to adequately predict the relationship between each exercise and specific knowledge concepts due to the ambiguous mapping between hidden states and exercises.

\textbf{Knowledge Tracing Based on Item Response Theory (IRT)}. Item response theory originates from the field of psychometrics and assumes that an underlying trait represents each candidate’s ability and can be observed through their response to items. The two models underlying the IRT model are the normal ogive model and the logistic model. However, the logistic regression model is the most common in practical applications \cite{baylor2011introduction}. The IRT model is based on four assumptions: 1) monotonicity (the probability of a correct response increases as the level of the trait increases). 2) one-dimensionality  (it is assumed that a dominant underlying trait is being measured). 3) local independence (responses to separate items in a given test are independent of each other at a given level of ability). 4) invariance (it is assumed that students’ abilities remain constant over time).

Here, ${\theta_i}$ is defined as the individual ability parameter of the $i-th$ student, ${a_j}$ is defined as the discrimination parameter of question $j$, ${b_j}$ is defined as the difficulty parameter of question $j$, and ${c_j}$ is defined as the guessing parameter of question $j$. With one-parameter IRT, it is possible to provide students with interpretable parameters in terms of two dimensions, personal ability and difficulty, as shown in Fig. \ref{fig:3-1-1-irt}. A two-parameter IRT model uses two parameters (difficulty and discrimination) to predict the probability of a successful response. Therefore, the discrimination parameter can vary between items and be plotted with different slopes, thus eliminating the explanatory information. A three-parameter IRT model adds a guessing parameter to the two-parameter model. The items answered by guessing indicate that the student’s ability is less than the difficulty of the question to which he or she is responding. The three-parameter IRT model can provide explanatory information about guessing behavior.

In IRT, the greater the difficulty of the item is, the greater the corresponding competencies needed by the student. By explicitly defining parameters such as item difficulty and student competencies, the model is transparent in its computational process and has good interpretability. However, the static IRT model assumes that students’ abilities remain constant over time, which is particularly unsuitable for long-term knowledge tracing.

\begin{figure*}[t]
    \centering
    \includegraphics[width=0.7\textwidth]{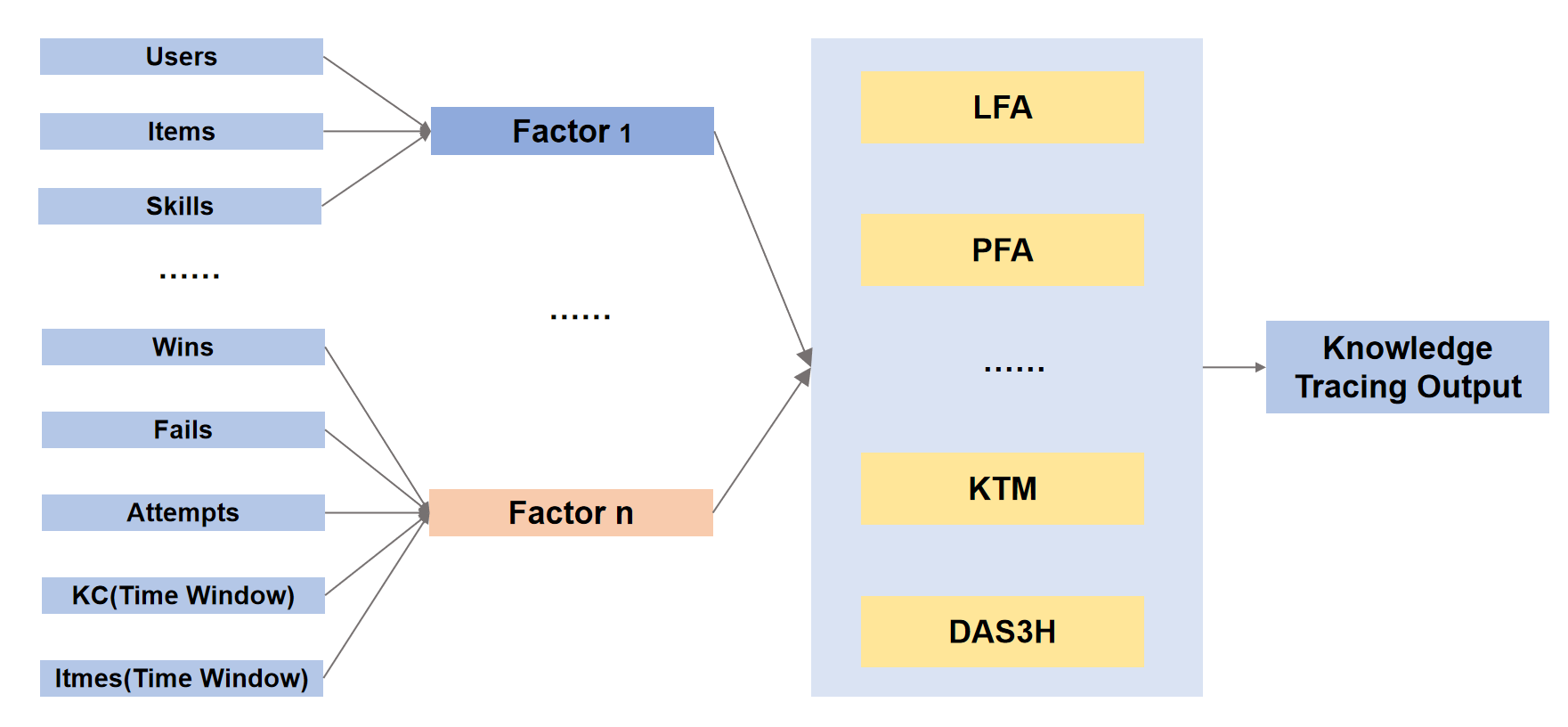}
    \caption{\textrm{Learning factor analysis approach.}}
    \label{LFA}
\end{figure*}

\textbf{Knowledge Tracing Based on Factor Analysis Model.}
Factor analysis is another critical approach for assessing learners’ knowledge mastery. Cen et al. \cite{cen2006learning} proposed learning factor analysis (LFA), a theory whose primary purpose is to find a more valid cognitive model from students’ learning data. Moreover, LFAs inherit the Q matrix used in psychometrics to assess cognition and extend the theory of learning curve analysis, as shown in Fig. \ref{LFA}.

LFA allows researchers to evaluate different representations of knowledge concepts by performing a heuristic search of the cognitive model space. Based on LFA theory, Cen et al. proposed the additive factor model (AFM) \cite{cen2008comparing} and performance factor analysis (PFA) \cite{pavlik2009performance}. The AFM is a particular case of PFA and is equivalent when ${\gamma _k}$ equals ${\rho _k}$. The AFM explains how the difficulty of a student’s knowledge points and the number of attempts to solve the problem affect the student’s performance, while PFA explains the student’s performance in terms of the difficulty of the knowledge points, the number of successes, and the number of failures.
\begin{table*}[!b]
    \caption{Overview of Factors Explained by Factor Analysis Models.}
    \label{tb-3-1-1-lfa-table}
    \resizebox{\textwidth}{!}{
    \scriptsize
    \begin{tblr}{
        colspec={Q[l,m,0.1\textwidth] Q[l,m,0.1\textwidth] Q[l,m,0.1\textwidth] Q[l,m,0.1\textwidth] Q[l,m,0.1\textwidth] Q[l,m,0.1\textwidth] Q[l,m,0.1\textwidth] Q[l,m,0.1\textwidth] Q[l,m,0.1\textwidth]},
        hspan=minimal,
    }
    \hline[1pt]
      Model         & Users               &Items              & Skills                 &Wins            &Fails         &Attempts        &KC(Time Window)           &Items(Time Window)        \\
    \hline
    IRT                             & \checkmark    & \checkmark    &        &      &       &          &        &            \\
    MIRT                            & \checkmark     & \checkmark     &        &      &       &          &        &            \\
    AFM                             &       &       & \checkmark     &      &       & \checkmark       &        &            \\
    PFA                             &       &       & \checkmark      & \checkmark   & \checkmark     &          &        &            \\
    KTM                             &       &       &        &      &       &          &        &            \\
    DASH                            & \checkmark    & \checkmark    &        & \checkmark   &       & \checkmark        &        & \checkmark          \\
    DAS3H                           & \checkmark     & \checkmark     & \checkmark      & \checkmark   &    & \checkmark    & \checkmark    &    \\
    Best-LR                         & \checkmark     & \checkmark     & \checkmark      & \checkmark    & \checkmark     &   & &    \\
    \hline[1pt]
    
    \end{tblr}}
\end{table*}

Large-scale factor analysis models have been further developed based on earlier factor analysis models. Using a factor decomposition approach, Vie et al. \cite{vie2019knowledge} proposed knowledge tracing machines (KTMs). KTMs use a sparse set of weights for all features to model the learner’s correct answer probability. The DASH (difficulty, ability, student interaction history) model is used for memory forgetting and factor analysis \cite{lindsey2014improving}. The DAS3H is a newer model that combines IRT and PFA and extends the DASH model by using a time window-based counting function to calculate characteristic factors \cite{lindsey2014improving,DBLP:conf/edm/ChoffinPBV19}. With the DAS3H model, the factor analysis method can explain student changes over a continuous time window, thus extending the scope of application of the factor analysis method. Gervet et al. \cite{gervet2020deep} proposed Best-LR based on DAS3H. Unlike DAS3H, Best-LR does not use a window but directly uses the number of successes or failures as an additivity factor. The factors that Best-LR can explain are similar to those that can explain DAS3H. The performance of Best-LR is better than that of DAS3H because Best-LR does not need to calculate window features.

In summary, logistic regression models can explain two main types of features: 1) coded embeddings representing questions and KCs and 2) counting-based features. In Table \ref{tb-3-1-1-lfa-table}, we compare the factors used by factor analysis models. Counting features summarize the history of students’ interactions with the system, and counting methods vary among different models, with some even introducing the concept of time windows.
\begin{figure*}[t]
    \centering
    \includegraphics[width=0.7\textwidth]{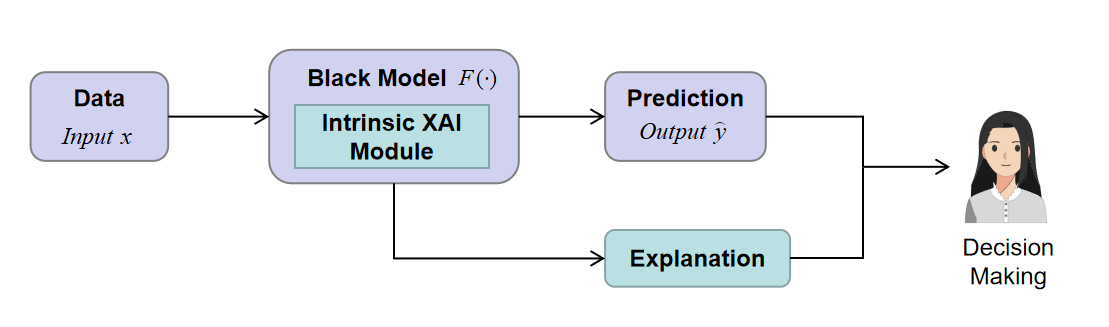}
    \caption{The framework of the model-intrinsic interpretability method.}
    \label{Ante-hoc}
\end{figure*}
\subsubsection{Category 2: Model-Intrinsic Interpretability}
Due to the invisibility of the internal structure and operation mechanism of the neural network model, model-intrinsic interpretability can only be realized by adding an interpretable module. Deep neural networks $F(\cdot)$ have large network layers and large parameter spaces. An end-to-end process is used to obtain the output prediction $\hat{y}$ from the input sample $x$. This process is similar to that of a black box. Therefore, researchers have attempted to embed a simple or an easy-to-interpret module inside the model to achieve model-intrinsic interpretability, thus resembling an interpretable model from the outside to provide explanations for the audience, as shown in Fig. \ref{Ante-hoc}. For instance, attention mechanisms can provide explanations by visualizing attention weights. As a result, attention mechanisms have become common intrinsic explainable modules in neural networks and are widely used in computer vision \cite{xu2015show,2021Transformer}, sentiment analysis \cite{2018Importance,2020Word}, recommendation systems \cite{wu2019dual,2022Interpretable}, and other fields.

In xAI, this kind of interpretation method is essentially explained by following strict axioms, rule-based final decisions, granular interpretations of decisions, etc. \cite{das2020opportunities}. It is worth noting that this method can only be used for a specific model, which leads to poor transferability. In the xKT model, as shown in Table \ref{KT1}, researchers have attempted to improve the model interpretability by introducing attention mechanisms, educational psychology, and other theories as interpretable modules. These model-intrinsic interpretability methods aim to make the model more transparent and understandable to stakeholders while maintaining good performance. In the following section, we elaborate on these two methods of model-intrinsic interpretability.

\begin{table*}[t]
    \caption{ Model-intrinsic interpretability for knowledge tracing.}
    \label{KT1}
    \resizebox{\textwidth}{!}{
    \footnotesize
    \begin{tblr}{
        colspec={Q[l,m,0.3\textwidth] Q[l,m,0.5\textwidth]},
        rowsep=0.6pt,
        hspan=minimal,
    }
    \hline[1pt]
      Model-intrinsic module         & References \\
    \hline
     Attention mechanism      &\cite{ghosh2020context} \cite{zhao2020interpretable} \cite{pandey2020rkt} \cite{zhang2021input} \cite{li2022knowledge} \cite{10191799}    \\\cline{2-2} 
     \SetCell[r=6]{l,m}{Educational psychology and other theories}     & Item response theory (IRT) \cite{DBLP:conf/edm/Yeung19} \cite{gan2020knowledge} \cite{converse2021incorporating} \cite{DBLP:conf/edm/Zhou0CZY021} \\
                                                & Multidimensional item response theory (MIRT) \cite{su2021time}\\
                                                & Constructive learning theory \cite{liu2022ability} \cite{sun2023progressive}\\
                                                & Learning curve theory and forgetting curve theory \cite{zhang2021learning}\\
                                                & Finite state automaton (FSA) \cite{zhu2020learning}\\
                                                & Classical test theory (CTT) \cite{lee2022monacobert}  \\
                                                & Monotonicity theory \cite{zhang2023counterfactual}   \\
                                                
    \hline[1pt]
    \end{tblr}
    }
\end{table*}

\textbf{Integrating attention mechanism module.} The self-attentive knowledge tracing (SAKT) model \cite{DBLP:conf/edm/PandeyK19} identifies concepts related to a given concept from historical student interaction data and predicts learners’ performance in the next exercise by considering related exercises in past interactions. This process involves sparse data. To address the problem that the attention layer is too shallow to recognize the complex relationships between exercises and responses, separated self-attentive neural knowledge tracing (SAINT) \cite{choi2020towards}, which is based on transformers and stacked two multihead attention layers on the decoder, was proposed to more effectively model the complex relationships between exercises and answers. The above work proved that the introduction of an attention mechanism into knowledge tracing greatly improves the performance of the model. Furthermore, several researchers have studied the construction of an attention mechanism for the knowledge tracing model as an interpretable module to improve the model’s explainability.

For example, Liu et al. \cite{liu2019ekt} proposed explainable exercise-aware knowledge tracing (EKT), which utilizes a novel attention mechanism to deeply capture the focusing information of students on historical exercises. This technique can track students’ knowledge states on multiple concepts and visualize knowledge acquisition tracing and student performance prediction to ensure the interpretability of the model. Context-aware attentive knowledge tracing (AKT) \cite{ghosh2020context} combines interpretable components into a novel monotonic attention mechanism and uses the Rasch model to regularize concepts and exercises; this approach has been proven to have excellent interpretability via experiments. Moreover, Zhao et al. \cite{zhao2020interpretable} proposed a novel personalized knowledge tracing framework with an attention mechanism that uses learner attributes to explain the prediction of mastery. The relation-aware self-attention model for knowledge tracing (RKT) \cite{pandey2020rkt} uses interpretable attention weights to help visualize the relationships between interactions and temporal patterns in the human learning process. Similarly, Zhang et al. \cite{zhang2021input} introduced a new vector to capture additional information and used attention weights to analyze the importance of input features, making it easier for readers to understand the predicted results. Recently, Li et al. \cite{li2022knowledge} regarded the attention mechanism as an effective interpretability module for constructing a new knowledge tracing model, effectively improving the interpretability and predictive ability of the model. Yue et al. \cite{yue2023augmenting} , based on ability attributes and an attention mechanism, provided explanations through an inference path. Zu et al. introduced CAKT\cite{zu2023cakt}, an innovative model that merges contrastive learning and attention networks to enable interpretable knowledge tracing.

Attention mechanisms, through visualized attention weights, explain aspects of decision-making in models. However, the interpretability of these tools is contingent upon the complexity of the model and the expertise of the interpreter. While elucidating certain decisions in simpler models, attention weights may become less transparent in more complex architectures, where multiple layers and nonlinear interactions obscure the interpretability of the information. Therefore, despite their utility, attention mechanisms should be integrated with complementary techniques for more holistic interpretability in sophisticated deep learning models.

\textbf{Integrating educational psychology and other theories.} 
Item response theory \cite{drasgow1990item} is a modern psychometric theory in which “items” refer to the questions in students’ papers and “item responses” refer to students’ answers to specific questions. As the parameters of IRT are interpretable, many scholars have combined IRT with deep learning methods, which have powerful feature extraction capabilities for enhancing interpretability. Deep-IRT \cite{DBLP:conf/edm/Yeung19} integrates dynamic key-value memory networks (DKVMNs) with IRT for knowledge training. The DKVMN captures learners’ trajectories, inferring their abilities and item difficulties via neural networks, which are subsequently utilized in IRT to predict answer correctness. This model combines the predictive strength of the DKVMN model with the interpretability of IRT, enhancing both the performance and insight into learner and item profiles.

Even though IRT can utilize predefined interpretable parameters to describe students’ behavior, students’ ability to solve problems is not limited; therefore, one-dimensional IRT parameters cannot be used to effectively explain students’ complex behaviors in a real-world scenario. To address this issue, enhanced deep multidimensional item response theory (TC-MIRT) \cite{su2021time} integrates the parameters of a multidimensional item response theory into an improved recurrent neural network, which enables the model to predict students’ states and generate interpretable parameters in each specific knowledge field. Inspired by the powerful interpretability of IRTs, many studies have integrated them into model frameworks to improve the model reliability in recent years. For example, knowledge interaction-enhanced knowledge tracing (KIKT) \cite{gan2020knowledge} uses the IRT framework to simulate learners’ performance and obtains an interpretable relationship between learners’ proficiency and project characteristics. Geoffrey Converse et al. \cite{converse2021incorporating} improved the model interpretability by transforming the representation of high-dimensional student ability from a deep learning model to an interpretable IRT representation at each time step; leveled attentive knowledge tracing (LANA) \cite{DBLP:conf/edm/Zhou0CZY021} uses the interpretable Rasch model to cluster students’ ability levels, thus using leveled learning to assign different encoders to different groups of students. Recently, Chen et al. \cite{DBLP:conf/aaai/00060HL023} developed QIKT, a question-centric KT model, improved knowledge tracing interpretability using question-centric representations and an interpretable item response theory layer.

In addition, constructivist learning theory \cite{fosnot2013constructivism} is a classical cognitive theory that emphasizes knowledge mastery differences as the result of knowledge internalization. Based on this theory, the ability boosted knowledge tracing (ABKT) model \cite{liu2022ability} utilizes continuous matrix factorization to simulate the knowledge internalization process for enhancing the model’s interpretability. PKT \cite{sun2023progressive} was designed based on constructivist learning and item response theories and features interpretable and educationally meaningful parameters. The forgetting curve theory \cite{averell2011form} indicates that a decrease in students’ memory during learning usually reduces their proficiency in knowledge concepts. The learning curve \cite{anzanello2011learning} regards knowledge acquisition as a mathematical expression in the process of human learning; that is, students can acquire knowledge after each practice. According to the above two types of pedagogical research, Zhang et al. \cite{zhang2021learning} constructed learning and forgetting factors at the learner level as additional factors to better trace and explain changes in learners’ knowledge levels. Moreover, some researchers have attempted to integrate a mathematical compression model into the KT model to enhance the interpretability of the model. For example, Wang et al. \cite{2023What} utilized finite state automation (FSA) to interpret the hidden state transition of DKT when receiving inputs. In addition, MonaCoBERT \cite{lee2022monacobert} uses a classical test theory-based (CTT-based) embedding strategy to consider the difficulty of an exercise to improve the performance and interpretability of the model. Recently, the counterfactual monotonic knowledge tracing (CMKT) \cite{zhang2023counterfactual} method enhances interpretability by integrating counterfactual reasoning with the emonotonicity theory in knowledge acquisition, demonstrating superior performance across real-world datasets.

Incorporating educational psychology theories into models offers interpretability through psychological frameworks and parameters. However, the efficacy of these methods in complex real-world educational contexts is limited and often constrained by the specificity and scope of the underlying theories. While providing insights into controlled scenarios, these approaches may struggle to encapsulate the multifaceted and dynamic nature of learning processes. Consequently, their application necessitates a nuanced and broadened perspective, blending theoretical insights with empirical data analysis to enhance the overall interpretability of the model in diverse educational environments.

\subsection{Stage 2: Post-hoc Interpretable Methods}

Post-hoc interpretability techniques are applied to pretrained machine learning models, especially those considered “black box” models, to explain their decisions, as shown in Fig. \ref{post-hoc}. Unlike ante-hoc methods, which are built into the model during development for inherent interpretability, post-hoc methods are employed after model creation, mainly to clarify the predictions \cite{adadi2018peeking,madsen2022post}. However, some post-hoc methods, such as knowledge distillation \cite{gou2021knowledge, wang2021knowledge}, rule extraction \cite{xue2022adaptive}, and activation maximization \cite{casarotto2022rt}, extend beyond explaining outputs; they also attempt to uncover the model’s internal mechanisms. In this paper, we delve into post-hoc interpretable methods for knowledge tracing from two angles: model-specific and model-agnostic methods. Model-specific methods are tailored to particular types of knowledge tracing models, reflecting their unique architectures and learning algorithms. Conversely, model-agnostic methods offer broader applicability, allowing for interpretability across various knowledge tracing models, regardless of their specific designs. This distinction is crucial for developing a comprehensive understanding of how different interpretability methods can be leveraged to demystify the predictions of knowledge tracing models, thereby enhancing their utility and trustworthiness in educational applications.

\begin{figure*}[t]
    \centering
    \includegraphics[width=0.8\textwidth]{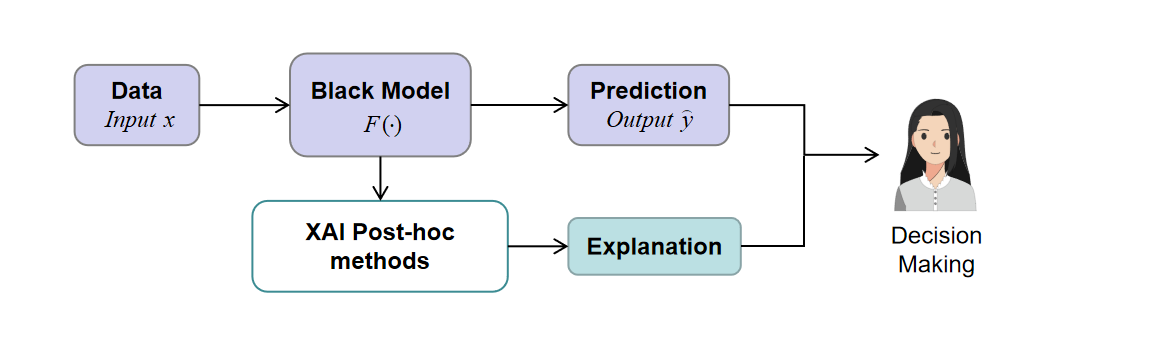}
    \caption{\textrm{The framework of the model-intrinsic interpretability method.}}
    \label{post-hoc}
\end{figure*}

\subsubsection{Category 1: Model-Specific}
Most model-specific methods focus on the interpretability of deep learning, mainly for a certain type of model. It is important to note that model-specific approaches are not necessarily model-based but specific to a class of models. At present, model-specific methods are mainly used to explain the following two categories of models: 1) ensemble-based models \cite{hara2018making,obregon2019rulecosi,konstantinov2021interpretable}; and 2) neural networks  \cite{craven2014learning,zeltner2021squashing}. Knowledge tracing based on deep learning is a black box that is difficult to understand due to the large parameter space inside the neural network. At present, researchers generally use visualization methods to explain this type of neural network in models.

\textbf{Layer-wise Relevance BackPropagation (LRP).}
The LRP technique was proposed by Bach et al. \cite{bach2015pixel} to calculate the correlation scores of single features in the input data by decomposing the output prediction of the deep neural network. It uses a specially designed set of propagation rules to operate a neural network by backpropagating predictions, where the inputs can be images, videos, or texts \cite{bach2015pixel,arras2017relevant}. In recurrent neural networks (RNNs), correlations are propagated to hidden states and memory units. Several researchers have applied the LRP method to the knowledge tracing model to enhance its interpretability. For example, Lu et al. \cite{lu2020towards} proposed solving the interpretability of deep DLKT by adopting the LRP method, which backpropagates the relevance from the output layer of the model to its input layer to explain the RNN-based DLKT model.  Wang et al. \cite{wang2021interpreting} studied whether the same post hoc interpretable method could be applied to the extensive dataset EdNet and achieved particular effectiveness. However, its effectiveness decreases with increasing learner size and learner practice sequence . Subsequently, Lu et al. \cite{lu2022interpreting} used the classical LRP method to interpret the output forecast variables of the DLKT model from the ready-made inputs of the DLKT model, which captured skill-level semantic information. The model output is progressively mapped to the input layer using a backward propagation mechanism, and the interpretation method assigns relevant values to each input sky-answer pair.

The LRP provides valuable insights into neural network-based KT models by mapping predictions back to input features. However, its reliance on correlation for attribution measurements raises concerns about fidelity, as it may be influenced by spurious correlations. This limitation, along with scalability issues in handling large datasets and complex learner sequences, as indicated in \cite{wang2021interpreting}, questions its applicability and relevance in diverse educational settings. The effectiveness of this method in treating KT thus requires careful consideration of these potential drawbacks.

\textbf{Causal Inference.} Causal inference is a method for analyzing causal relationships in observational data, attempting to determine whether different treatments (such as different strategies or methods in an experiment) lead to different outcomes \cite{yao2021survey,pearl2010causal}. The focus is on distinguishing true causal effects from mere correlations, especially when dealing with confounding variables \cite{dinga2020controlling}. Causal inference enhances the transparency and interpretability of AI models by clarifying the “why” behind AI decisions and distinguishing between direct causal relationships and spurious associations. Zhu et al.\cite{zhu2023stable} focused on applying causal inference to the field of knowledge tracing. By adjusting confounding variables within a causal inference framework, they aimed to enhance the prediction accuracy and stability of knowledge tracing models. This approach takes into account key factors, such as confounding variables, to improve the models' ability to predict students' knowledge states accurately. Furthermore, the temporal and causal-enhanced knowledge tracing (TCKT) model\cite{huang2024learning} integrates causal self-attention with temporal dynamics. This integration not only enhances prediction accuracy and interpretability in educational settings but also effectively mitigates dataset bias by employing causal inference to model the student learning process more accurately.

Causal inference, which is critical in distinguishing between correlation and causation, is invaluable in KT for analyzing the impact of educational interventions. The challenges, as outlined in \cite{yao2021survey}, lie in the need for robust statistical frameworks and the management of confounding variables, which can be daunting in practical educational contexts. Its application in KT requires a careful balance between theoretical robustness and practical feasibility.

\textbf{Visualization.} A classic approach to interpret black box models is visualization, which provides intuitive explanations through analysis of the model’s training process. Based on their previous work \cite{ding2019deep}, Ding et al. \cite{ding2021interpretability} tried to open the “box” of the deep knowledge tracing model. First, they used the larger dataset EdNet to visually analyze the behavior of the DKT model in high-dimensional space, tracked the changes in activation over time, and analyzed the influence of each skill relative to other skills, which solved the problem that interpretation methods were not intuitive.

Visualization techniques provide an intuitive means of interpreting complex KT models. However, they necessitate a high level of expertise in both the model’s workings and the data represented. The risk here, particularly with high-dimensional data, is the potential for oversimplification or misinterpretation of the model’s dynamics, leading to incorrect conclusions about the learning process.

\textbf{Ensembling Approaches.} Several researchers have attempted to use ensemble approaches to improve the interpretability of knowledge tracing models on big data. For example, Tirth Shah et al. \cite{shah2020explainable} used a combination of 22 models to predict whether students can answer given questions correctly and discovered that an ensembling approach can improve the prediction performance and interpretability of knowledge tracing tasks. EnKT \cite{sun2022ensemble} is based on BKT and DKT and represents student concepts and student questions using learning and performance parameters, respectively, to improve the interpretability of the model.

Ensembling approaches combine multiple models to enhance both predictive accuracy and interpretability in KT. However, the increased complexity of these methods can obscure the contributions of individual models within the ensemble. This complexity poses a significant challenge in KT, where understanding the specific influence of different factors on learning outcomes is crucial.

\clearpage
{
    \centering
    \footnotesize
    \refstepcounter{table}
    \DefTblrTemplate{contfoot-text}{normal}{\textit{Continued on the next page}}
    \SetTblrTemplate{contfoot-text}{normal}
    \begin{longtblr}[
    caption = {A summary of different types of explainable knowledge tracing models.},
    label = {summary-table}
    ]{
        colspec={Q[l,m,0.15\textwidth] Q[c,m,0.05\textwidth] Q[c,m,0.08\textwidth] Q[c,m,0.08\textwidth] Q[c,m,0.08\textwidth] Q[c,m,0.08\textwidth] Q[c,m,0.06\textwidth] Q[c,m,0.065\textwidth] Q[c,m,0.06\textwidth]},
        rowsep=0.6pt, 
        hspan=minimal,
        vline{2,3}={1-Z}{0.5pt},vline{5}={1-Z}{0.5pt},vline{7}={2-Z}{0.5pt},vline{9}={2-Z}{0.5pt}
    }
    \hline[1pt]
      \SetCell[r=3]{c,m}{Models}    & \SetCell[r=3]{c,m}{Year}    & \SetCell[c=2]{c,m}{Taxonomy} &   & \SetCell[c=5]{c,m}{Interpretable Methods}     &  &    &    &    \\ \cline{3-9} 
                                    &           & \SetCell[r=2]{c,m}{Transparent Model} & \SetCell[r=2]{c,m}{Black-box model} & \SetCell[c=2]{c,m}{Ante-hoc} &  & \SetCell[c=2]{c,m}{Post-hoc} &  &\SetCell[r=2,c=1]{c,m}{Other Dimensions}  \\ \cline{5-9} 
                                    &         &     &       & Self-explanatory & Model-Intrinsic  & Model-Specfic & Model-Agnostic    &      \\                             
    \hline
    BKT \cite{Albert1994Knowledge}           & 1994         & \checkmark      &         & \checkmark       &          &          &           &     \\ 
    Pardos et al. \cite{pardos2010modeling}  & 2010         & \checkmark      &         & \checkmark       &          &          &           &     \\ 
    Lee et al. \cite{lee2012impact}          & 2012         & \checkmark      &         & \checkmark       &          &          &           &     \\ 
    BKT-ST \cite{hawkins2014using}   & 2014         & \checkmark      &         & \checkmark       &          &          &           &     \\ 
    Wang et al. \cite{wang2016structured}    & 2016         & \checkmark      &         & \checkmark       &          &          &           &     \\ 
    Sun et al. \cite{sun2022genetic}         & 2022         & \checkmark      &         & \checkmark       &          &          &           &     \\ 
    IKT \cite{minn2022interpretable}
   &2022         & \checkmark               &       &             &          &          &           & \\
    Affective BKT \cite{spaulding2015affect}   & 2015         & \checkmark      &         & \checkmark       &          &          &           &     \\ 
    PC-BKT \cite{nedungadi2014predicting}& 2014         & \checkmark      &         & \checkmark       &          &          &           &     \\ 
    LFA \cite{cen2006learning}                & 2006         & \checkmark      &         & \checkmark       &          &          &           &     \\  
    AFM \cite{cen2008comparing}               & 2008         & \checkmark      &         & \checkmark       &          &          &           &     \\
    PFA \cite{pavlik2009performance}                & 2009         & \checkmark      &         & \checkmark       &          &          &           &     \\
    KTM \cite{vie2019knowledge}               & 2019         & \checkmark      &         & \checkmark       &          &          &           &     \\
    DASH \cite{lindsey2014improving}          & 2014         & \checkmark      &         & \checkmark       &          &          &           &     \\
    DAS3H \cite{DBLP:conf/edm/ChoffinPBV19}             & 2019     & \checkmark      &         & \checkmark       &          &          &           &     \\
    Best-LR \cite{gervet2020deep}           & 2020         & \checkmark      &         & \checkmark       &          &          &           &     \\
    AKT \cite{ghosh2020context}              & 2020         &            & \checkmark         &             &\checkmark     &          &           &     \\ 
    RKT \cite{pandey2020rkt}                 & 2020         &            & \checkmark         &             &\checkmark     &          &           &     \\ 
    Zhang et al. \cite{zhang2021input}       & 2021         &            & \checkmark         &             &\checkmark     &          &           &     \\ 
    Li et al. \cite{li2022knowledge}         & 2022         &            & \checkmark         &             &\checkmark     &          &           &     \\ 
    Yue et al. \cite{yue2023augmenting}         & 2023         &            & \checkmark         &             &\checkmark     &          &           &     \\ 
    CAKT \cite{zu2023cakt}         & 2023         &            & \checkmark         &             &\checkmark     &          &           &     \\     
    MonaCoBERT \cite{lee2022monacobert}      & 2022         &            & \checkmark         &             &\checkmark     &          &           &     \\
    CMKT \cite{zhang2023counterfactual}             & 2023         &            & \checkmark         &             &\checkmark     &          &           &     \\
    deep-IRT \cite{DBLP:conf/edm/Yeung19}            & 2019         &            & \checkmark         &             &\checkmark     &          &           &     \\
    TC-MIRT \cite{su2021time}                & 2011         &            & \checkmark         &             &\checkmark     &          &           &     \\
    KIKT \cite{gan2020knowledge}             & 2020         &            & \checkmark         &             &\checkmark     &          &           &     \\
    Geoffrey Converse et al. \cite{converse2021incorporating}
    & 2021         &            & \checkmark         &             &\checkmark     &          &           &     \\
    LANA \cite{DBLP:conf/edm/Zhou0CZY021}
    & 2021         &            & \checkmark         &             &\checkmark     &          &           &     \\
    QIKT \cite{DBLP:conf/aaai/00060HL023}
    & 2023         &            & \checkmark         &             &\checkmark     &          &           &     \\
    ABKT \cite{liu2022ability}
    & 2022          &            & \checkmark         &             &\checkmark     &          &           &     \\
    PKT\cite{sun2023progressive} 
    & 2023          &            & \checkmark         &             &\checkmark     &          &           &     \\
    zhang et al. \cite{zhang2021learning}
    &2021          &            & \checkmark         &             &\checkmark     &          &           &     \\
    EAKT \cite{pu2022embedding}
    &2022          &            & \checkmark         &             &\checkmark     &          &           &     \\
    Zhu et al. \cite{zhu2020learning}
    &2022          &            & \checkmark         &             &\checkmark     &          &           &     \\
    Ding et al. \cite{ding2019deep}
    &2019     &            & \checkmark         &             &          &\checkmark     &           &     \\
    Ding et al. \cite{ding2021interpretability}
    &2021     &            & \checkmark         &             &          &\checkmark     &           &     \\
    Zhu et al.\cite{zhu2023stable}
    &2023     &            & \checkmark         &             &          &\checkmark     &           &     \\
    TCKT \cite{huang2024learning}
    &2024     &            & \checkmark         &             &          &\checkmark     &           &     \\
    Tirth Shah et al. \cite{shah2020explainable}
    &2020          &            & \checkmark         &             &          &\checkmark     &           &     \\
    EnKT \cite{sun2022ensemble}
    &2022          &            & \checkmark         &             &          &\checkmark     &           &     \\
    Lu et al. \cite{lu2020towards}
    &2020          &            & \checkmark         &             &          &          &\checkmark      &     \\
    Valero et al. \cite{valero2023shap}
    &2023          &            & \checkmark         &             &          &          &\checkmark      &     \\ 
    Wang et al. \cite{wang2021interpreting}
    &2021          &            & \checkmark         &             &          &          &\checkmark      &     \\
    Lu et al. \cite{lu2022interpreting}
    &2022          &            & \checkmark         &             &          &          &\checkmark      &     \\
    Varun Mandalapu et al. \cite{mandalapu2021we} 
    &2021          &            & \checkmark         &             &          &          &\checkmark      &     \\
    Wang et al. \cite{wang2022generic}
    &2022         &             & \checkmark         &             &          &          &\checkmark      &     \\
    GCE \cite{li2023genetic}
    &2023         &             & \checkmark         &             &          &          &\checkmark      &     \\
    Liu et al. \cite{liu2019ekt}
    &2019         &             & \checkmark         &             &          &          &           & \checkmark \\
    SPDKT \cite{dai2021improved}
    &2021         &             & \checkmark         &             &          &          &           & \checkmark \\
    CoKT \cite{sun2021collaborative}
    &2021         &             & \checkmark         &             &          &          &           & \checkmark \\
    Lee et al. \cite{lee2019knowledge}
    &2019         &             & \checkmark         &             &          &          &           & \checkmark \\
    HGKT \cite{tong2022introducing}
    &2022         &             & \checkmark         &             &          &          &           & \checkmark \\
    GKT \cite{nakagawa2019graph}
    &2019         &             & \checkmark         &             &          &          &           & \checkmark \\
    SKT \cite{tong2020structure}
    &2020         &             & \checkmark         &             &          &          &           & \checkmark \\
    Zhao et al.\cite{zhao2022research}
    &2022         &             & \checkmark         &             &          &          &           & \checkmark \\
    JKT \cite{song2021jkt}
    &2021         &             & \checkmark         &             &          &          &           & \checkmark \\
    
    \hline[1pt]
    \end{longtblr}  
    
\addtocounter{table}{-1}
}

\subsubsection{Category 2: Model-Agnostic}
Model-agnostic techniques separate explanations from model outputs and are applicable to any machine learning model \cite{adadi2018peeking}. It only acts on the input and output of the neural network, providing explanations by perturbing the input or simplifying the model. Because model-agnostic techniques are not limited to a specific model, most researchers currently prefer model-agnostic approaches over model-specific approaches.

\textbf{Local Interpretable Model-Agnostic Explanations (LIME).} The LIME was proposed by Ribeiro et al. \cite{ribeiro2016should}. This method trains local surrogate models to explain a single prediction a global black-box model gives. LIME partially replaces complex models with simpler models to provide local explanations. Specifically, since the perturbed data will affect the model’s output, LIME trains a local interpretable model to learn the mapping relationship between the perturbed data and the model’s output and uses the similarity between the perturbed input and the original input as the weight. Finally, the essential K features are selected from the local interpretable model for interpretation. This approach can provide a very effective local approximation to the black box model. In xKT, Varun Mandalapu et al. \cite{mandalapu2021we} utilized LIME to understand the impact of various features on best-performing model predictions.

LIME provides microlevel insights into specific predictions of KT models by training local interpretable surrogate models. Its major strength lies in revealing the influence of particular features in specific instances. However, LIME’s focus on local explanations may not capture the model’s global behavior, particularly in KT, where diverse learning paths can significantly influence model decisions. Additionally, the dependence of LIME on perturbation strategies and the choice of local models might affect the consistency and accuracy of its interpretations.

\textbf{Deep SHapley Additive exPlanations (Deep SHAP).} By combining DeepLIFT \cite{shrikumar2017learning} with Shapley values \cite{shapley1997value}, Lundberg and Lee \cite{lundberg2017unified} proposed a fast method to approximate Shapley values for CNNs called Deep SHAP. Deep SHAP decomposes the prediction of the deep learning model into the sum of feature contributions through backpropagation and obtains the reference specific contribution of each feature to the prediction through the backpropagation prediction difference. Several studies \cite{kim2021student} have concentrated on using the DKT model to predict test scores based on skill mastery and then assessed the influence of each skill on the predicted score using SHAP analysis. Inspired by this idea, Valero-Lea et al. \cite{valero2023shap} aimed to explain learners’ skill mastery by analyzing past interactions using a SHAP-like method to determine the importance of these interactions. Wang et al. \cite{wang2022generic} proposed a four-step procedure to interpret the DLKT model and obtained effective explicable results. The four-step procedure is as follows: 1) Given a sample x to be interpreted, reference samples are selected, and predictions are made about the last questions; 2) the difference between each reference sample and sample x is calculated; 3) the prediction is then backpropagated from the output layer to the input layer to calculate the reference-specific feature contribution between each reference sample and the interpreted sample; and 4) the contribution of each question-answer in sample x to the prediction of the DLKT model is obtained.

Deep SHAP, which combines DeepLIFT with Shapley values, elucidates feature contributions to predictions in deep learning models. KT helps us understand the relative importance of various features, such as prior performance or interaction frequencies. While effective at revealing individual feature impacts, Deep SHAP may struggle with high-dimensional feature spaces and overlook complex interfeature interactions, which are crucial in KT with diverse learning trajectories.
 
\textbf{Causal Explanation(CE).}
Li et al. \cite{li2023genetic} addressed the explainability issue of DLKT models by proposing a genetic causal explainer (GCE) based on genetic algorithms (GAs). The GCE established a causal framework and a specialized coding system, effectively resolving the issues of spurious correlations caused by reliance on gradients or attention scores, thereby enhancing the accuracy and readability of explanations. Additionally, the GCE is a post hoc explanation method that can be applied to various DLKT models without interfering with model training, offering a flexible and effective means of explanation.

GCE based on genetic algorithms addresses explainability in DLKT models by establishing a causal framework. While GCE offers a novel approach to understanding deep causal relationships in KT, establishing causal connections requires precise data modeling and hypothesis validation. The complexity and computational demands of GCE, particularly for large datasets, pose significant challenges.

\subsection{Stage 3: Other Dimensions}
Beyond the mainstream ante-hoc and post-hoc methods, there are interpretatable approaches specific to the knowledge tracing domain, yet underrepresented in current xAI literature. These approaches exploit unique data features in KT, such as interconnected relationships among questions, concepts, or users. Upcoming sections will provide a detailed exploration of these approaches and their role in enhancing interpretability.

\subsubsection{Category 1: Output Format}
\textbf{Visual explanations.} To verify the interpretability of the proposed model, many researchers use radar charts or heat maps to provide readers with visual explanations. In the following, several representative works will be introduced. Self-paced deep knowledge tracing (SPDKT) \cite{dai2021improved} reflects the difficulty of the problem by assigning different weights to the problem, visualizing the difficulty of the problem, and improving the interpretability of the model. Similarly, collaborative embedding method for Knowledge Tracing (CoKT) \cite{sun2021collaborative} provided an interpretable question embedding by visualizing the distance between question embeddings that share the same concepts and those that do not. Lee et al. \cite{lee2019knowledge} proposed a knowledge query network (KQN) model, which uses the dot product between the knowledge state vector and skill vector to define knowledge interaction and uses a neural network to encode students’ responses and skills into vectors of the same dimension. Moreover, the KQN can query students’ knowledge of different skills and subsequently enhance the interpretability of the model by visualizing the interaction between two types of vectors.

Visual explanations enhance surface-level interpretability in KT but often fail to delve into the internal mechanisms of models. The explanatory power of these models heavily relies on the quality of the data representation; inaccurate representations may lead to misleading interpretations. Moreover, complex visual outputs, such as relationships in high-dimensional spaces, can be challenging for general users to understand, limiting their practical effectiveness. Therefore, these methods require further refinement and development to provide more in-depth and thorough explanations.

\subsubsection{Category 2: Data Attribute}
\textbf{Interaction between exercises and concepts.} Several researchers have shown that the relationship between exercises and concepts may be structured into a graph by analyzing learners’ learning data. For example, one exercise may involve multiple concepts, and one concept may also correspond to multiple exercises. In addition, two relationships exist among concepts, prerequisite and similarity, as shown in Fig. \ref{rel}. The prerequisite is that the mastery of concept $y$ requires the mastery of concept $x$, for example, by adding before multiplying; the similarity relationship means that knowledge $y$ and knowledge $x$ belong to the same category, such as addition. Some studies show that incorporating this graph structure into the knowledge tracing model as a relation induction deviation can enhance interpretability in the prediction process. Generally, there are exercises-to-exercises, concepts-to-concepts, exercises-to-concepts, and other potential relationships in the graph structure constructed from learner learning data. Next, this subsection describes in detail the works that have been done to increase the interpretability of the model by using the underlying graph structure in the data.

\begin{figure*}[ht]
    \centering
    \includegraphics[width=\textwidth]{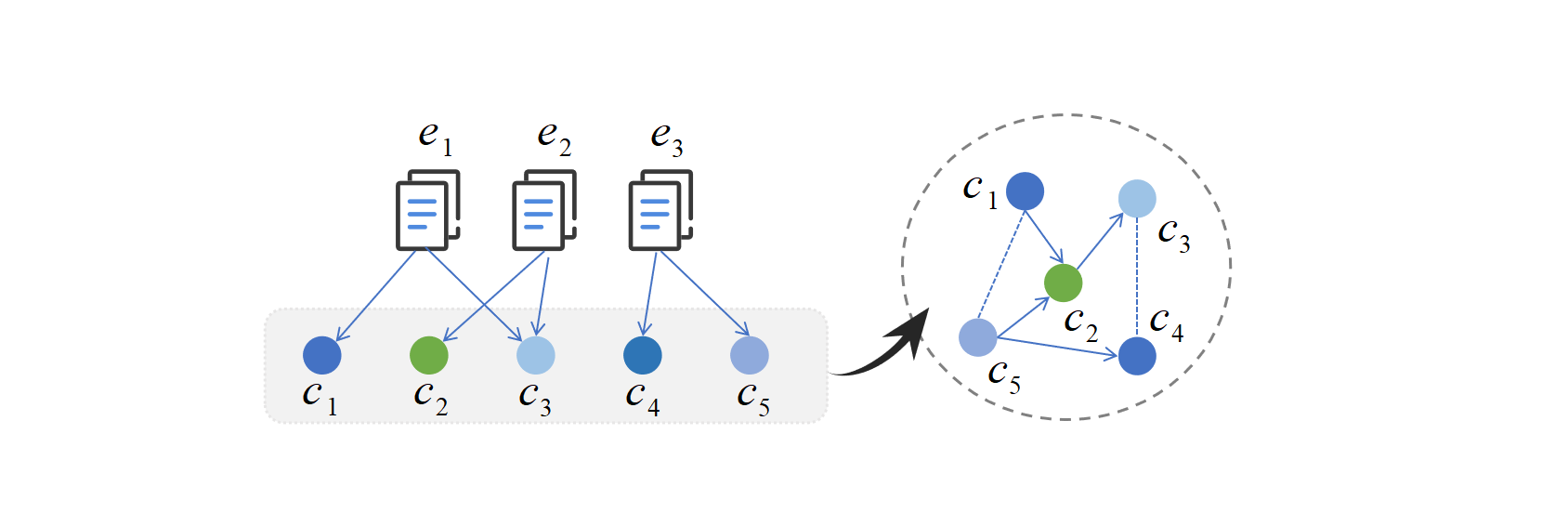}
    \caption{\textrm{ (a) Relationships between the exercises and the concepts. (b) Relationships between concepts: the solid line represents the prerequisite relationship, and the dashed line represents the similarity relationship.}}
    \label{rel}
\end{figure*}
At the level of exercise-to-exercise relationships, hierarchical graph knowledge tracing (HGKT) \cite{tong2022introducing} constructs a hierarchical exercise graph according to the latent hierarchical relationships (direct relationships and indirect relationships) between exercises and introduces a problem schema to explore the dependencies of exercise learning, which enhances the interpretability of knowledge tracing. On the level of concept-to-concept relationships, the graph-based knowledge tracing model (GKT) \cite{nakagawa2019graph} synchronously learns the latent relationships between concepts in the prediction process. The model updates the knowledge state related to the current exercises each time, thus realizing interpretability at the concept level. Educational entities emphasize the significance of knowledge structure. Structure-based knowledge tracing (SKT) \cite{tong2020structure} utilizes two different propagation models to track the influence of prerequisites or similarity relationships between concepts and has been used in many experiments to prove interpretability. This work visually demonstrated the interpretability of the models in the manner described in 3.1, such as through heatmaps and radar maps. On the level of exercise-to-concept relationships, Zhao et al. \cite{zhao2022research} used a graph attention network to learn the underlying graph structure between concepts in the answer record and input information from the model containing the relationship information between the exercises and the concept, which enhances the interpretability of the model. To dig deeper into the relationship between exercise-to-exercise and concept-to-concept, a joint graph convolutional network-based deep knowledge tracing (JKT) \cite{song2021jkt} framework was used to model the multidimensional relationships of the above two factors into a graph and fuse them with “exercise-to-concept” relationships. The model connects exercises under cross-concepts and helps capture high-level semantic information, which increases the interpretability of the model.

Graph-based approaches in knowledge tracing offer intricate insights into the relationships among exercises, concepts, and hierarchical interdependencies. While these methods enhance interpretability by mapping complex educational theories onto graph structures, they also present challenges in terms of complexity and accessibility. Their reliance on sophisticated graph representations and computational models may limit usability for nontechnical users such as teachers and students, hindering their practical application in diverse educational settings. Moreover, the assumptions inherent in these graph-based models about learning processes and relationships might not fully align with the dynamic and varied nature of real-world learning, raising questions about their generalizability and effectiveness.

\subsection{Explainable Knowledge Tracing: Application}
In this section, we explore the practical application of xKT in educational settings. Our focus is on its role in generating diagnostic reports, personalized learning, resource recommendations, and knowledge structure discovery. This section examines how xKT algorithms are used to track and predict learners' knowledge states, enabling the creation of dynamic, personalized educational pathways. We discuss the balance between algorithmic complexity and the need for clear, interpretable results in educational settings.

{\textbf{Knowledge tracing in diagnostic reports and visualization.}
In the realm of educational technology, knowledge tracing primarily manifests in the generation and visualization of learning diagnostic reports. Algorithms such as BKT and DLKT have been pivotal in this regard. BKT utilizes probabilistic modeling to continually update a student’s knowledge state, adjusting the likelihood of concept mastery after each educational interaction. DLKT, leveraging neural network architectures, excels in capturing complex learning patterns, offering nuanced insights into student performance. Despite its interpretability, BKT sometimes struggles with complex learning scenarios, whereas DLKT, though proficient at deciphering intricate behaviors, compromises clarity for the sake of complexity.

To enhance the interpretability of diagnostic reports, models such as KSGKT \cite{gan2022knowledge} integrate knowledge structures with graph representations, employing attention mechanisms to accurately predict learner performance. HGKT \cite{tong2022introducing} uses a hierarchical graph neural network to analyze learner interactions, enabling detailed categorization of exercises for a deeper understanding of learner knowledge and problem-solving skills. Both models aim to provide granular diagnostic reports to support personalized learning paths. Additionally, visual explanations, such as intuitive graphs and heatmaps \cite{dai2021improved}, significantly augment the interpretability of KT algorithm outputs, transforming complex data into actionable insights for tailored educational strategies.

\textbf{Knowledge Tracing in Personalized Learning and Resource Recommendation.}} The field of personalized learning and resource recommendation has greatly benefited from advancements in knowledge tracing algorithms. These algorithms are adept at tailoring learning pathways and recommending appropriate learning resources based on a student’s current state of knowledge and learning preferences \cite{munoz2022systematic}. The introduction of deep learning technologies, such as DKT, has further enhanced the precision and personalization of these recommendations. However, one limitation is the often reduced explainability of these sophisticated models, which can make it challenging for educators to understand the rationale behind specific recommendations.

Several studies have made notable efforts to enhance the accuracy and interpretability of personalized recommendations. The integration of concept tags with the DKVMN model, as seen in \cite{DBLP:conf/edm/AiCGZWFW19}, marks a significant step in improving exercise recommendations by accurately tracing students’ knowledge states. Building upon this, Zhao et al. \cite{10.1145/3386527.3406739} introduce attention mechanisms and learner attributes to refine mastery predictions, yielding more personalized and interpretable activity recommendations. Adopting a dynamic approach, Cai et al. \cite{INSPEC:19530087} combined reinforcement learning with knowledge tracing, adapting learning paths in real time to align with the learner’s evolving understanding. Similarly, the ER-KTCP \cite{DBLP:journals/ccftpci/HeWPZS22} innovatively merges knowledge state tracking with concept prerequisites for exercise selection, demonstrating marked improvements in student performance. Furthermore, Wang et al. \cite{wan2023pedagogical} focused on small private online courses (SPOCs), employing learning behavior dashboards and a modified DKVMN model to emphasize student engagement and concept mastery in a specific educational setting. Collectively, these studies contribute to a more nuanced understanding of data-driven models in educational technology, paving the way for adaptive, personalized learning experiences.

\textbf{Knowledge Tracing in Knowledge Structure Discovery.}
In knowledge structure discovery, knowledge tracing algorithms clarify the relationships between problems and concepts. They analyze students’ learning behaviors and performances to identify connections, aiding educators in understanding the foundational concepts for advanced problems. For example, an algorithm can show that mastering basic mathematical skills is essential before complex concepts are grasped. These insights are vital for creating effective teaching strategies and curricula, allowing educators to logically sequence lessons and ensuring that students master fundamentals before progressing to advanced topics.

Advancements such as HGKT \cite{tong2022introducing} and GKT \cite{nakagawa2019graph} have significantly improved the interpretability of learning models. HGKT reveals complex interdependencies between exercises, enhancing the interpretability of exercise-related learning progress. Moreover, GKT delves into the latent relationships between concepts, providing clear insights at the concept level. Complementary approaches such as SKT \cite{tong2020structure} and graph attention networks further augment this clarity by tracing relationships (both pre-requisite and similarity) between concepts. These methods, along with the JKT \cite{song2021jkt}, collectively enhance the overall interpretability of knowledge structures, making the connections within the learning process more understandable and accessible.

\subsection{Discussion}
In this comprehensive study, we have evaluated various xKT models, emphasizing interpretability, accuracy, computational efficiency, and applicability in real-world scenarios. This analysis elucidates the distinct characteristics and constraints of different xKT models, pivotal for enhancing educational technology tools.

\textbf{Interpretability.} Within KT models, the spectrum of interpretability ranges from transparent, ante hoc methods to intricate, post hoc techniques. Transparent models such as BKT offer straightforward interpretability due to their simple probabilistic frameworks, which are beneficial in scenarios demanding clarity \cite{Albert1994Knowledge}. In contrast, post hoc methods such as the LRP \cite{lu2022interpreting} and LIME \cite{mandalapu2021we} methods provide insights into more complex models suitable for detailed analytical requirements. However, these methods can be challenging for nontechnical users to interpret due to their complexity.

\textbf{Accuracy.} DLKT models, such as those employing neural networks, exhibit high accuracy in modeling complex student interactions but require substantial tuning and expertise \cite{ghosh2020context,su2021time}. These models, while powerful, can be prone to overfitting and opaque, making their predictions difficult to interpret. On the other hand, simpler models such as the BKT, despite being more interpretable, may not capture complex learning behaviors effectively, thus limiting their accuracy in more nuanced educational settings.

\textbf{Computational Efficiency.} The computational demands of KT models vary widely. Simpler structures such as those in BKT models are computationally efficient and align well with resource-constrained environments. Advanced models, particularly ensembling approaches \cite{sun2022ensemble}, demand significant computational power, making them suitable for well-resourced scenarios but impractical for more constrained scenarios.

\textbf{Real-world Applicability.} The applicability of a KT model heavily depends on the educational context. Transparent models \cite{spaulding2015affect} are ideal in settings where quick and clear feedback is essential. In contrast, environments that require deep insights into intricate learning patterns require more sophisticated models. However, these advanced models, while offering detailed analysis capabilities, often involve the downside of higher computational requirements and potential overfitting issues \cite{liu2022ability}.

In conclusion, the choice of a xKT model requires a delicate balance between interpretability, educational environment complexity, accuracy needs, and computational resource availability. Future developments in xKT should aim to integrate these factors, pursuing models that provide both clarity and depth in understanding diverse learning patterns adaptable across various educational contexts.

\section{Explainable Knowledge Tracing: Evaluation Method}\label{sub4}
Although many researchers have shown interest in explainable knowledge tracing, most model evaluation metrics focus on evaluating the model performance, while existing research evaluating interpretability is scarce \cite{song2022survey}. The process of imparting knowledge from teachers to learners needs to be highly explanatory and understandable. As an intelligent auxiliary tool in the teaching process, an AI model cannot be trusted by users only by its high accuracy \cite{fiok2022explainable,liu2022trustworthy}. Based on the above, it is also worth evaluating these existing explainable knowledge tracing models.

The following sections begin with a brief introduction to the common evaluation methods in xAI and then explore how we can develop a standardized and reasonable interpretable evaluation system for educational models on knowledge training tasks. The goal is to improve the user’s understanding and trust in an education model and realize the wide application of intelligent education products in education.

\subsection{A Case Study: Comparison of Interpretable Methods for Knowledge Tracing.}
As mentioned in Section \ref{sub3}, transparent models such as Bayesian knowledge tracing are explained by their internal parameters, while deep learning-based knowledge tracing requires additional specific interpretation methods. "Are all models in all defined-to-be-interpretable model classes equally interpretable \cite{doshi2017towards}?" To compare the unified interpretation results for the same model and dataset, we use three common post hoc xAI interpretable methods, LRP, LIME, and SHAP, to explain the deep knowledge tracing in ASSISTment2009 \cite{feng2009addressing}, as shown in Fig. \ref{compare}. For this dataset, we selected two interactive sequences, each with a length of 25, from two different learners to be explained. Using the aforementioned methods, we calculated and visualized the interpretable features of the interactive sequences for each respondent. This comparison provides insights into the effectiveness and interpretability of each method, helping us to better understand how the model can make predictions for deep knowledge tracing.

\begin{figure*}[!htb]
    \centering
    \includegraphics[width=0.9\textwidth]{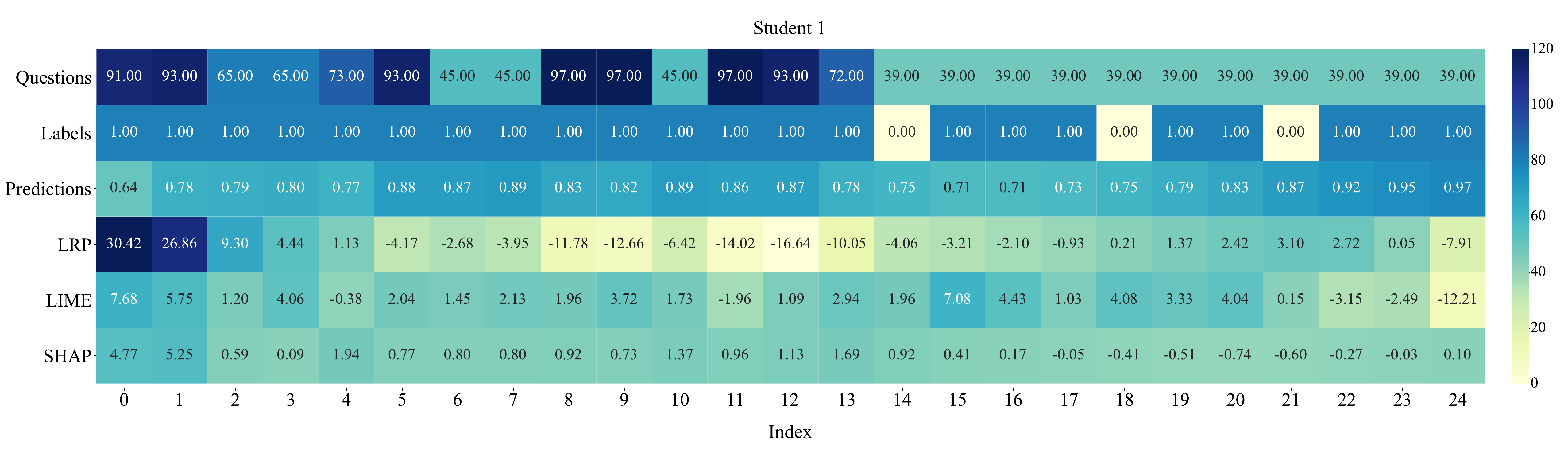}
     \includegraphics[width=0.9\textwidth]{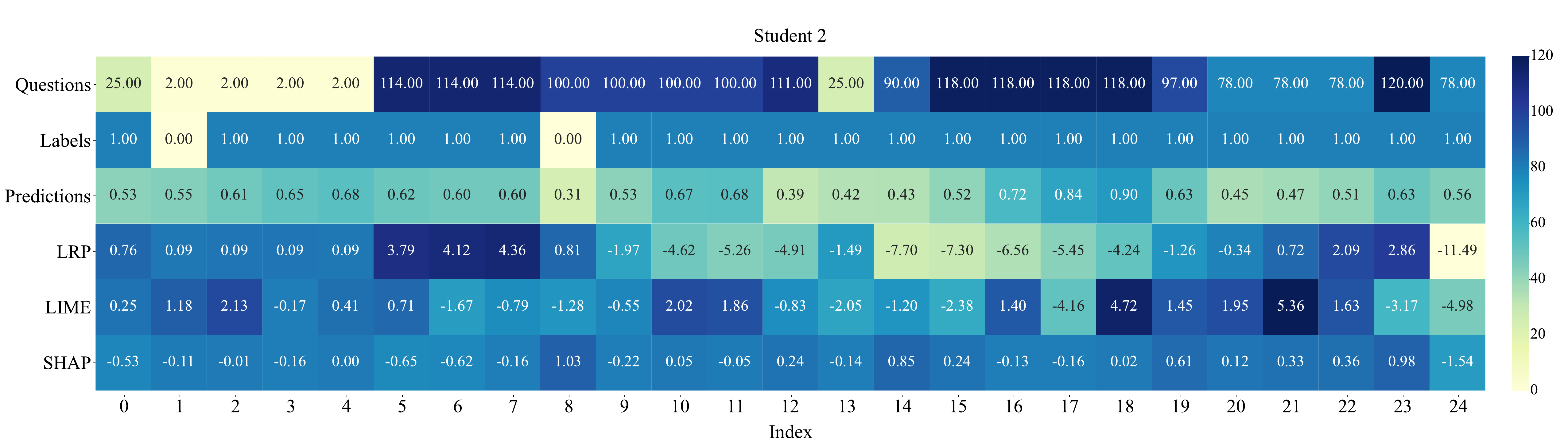}
    \caption{\textrm{Comparison of post-hoc interpretability methods on deep knowledge tracing.}}
    \label{compare}
\end{figure*}

\begin{figure*}[!htb]
    \centering
    \includegraphics[width=0.7\textwidth]{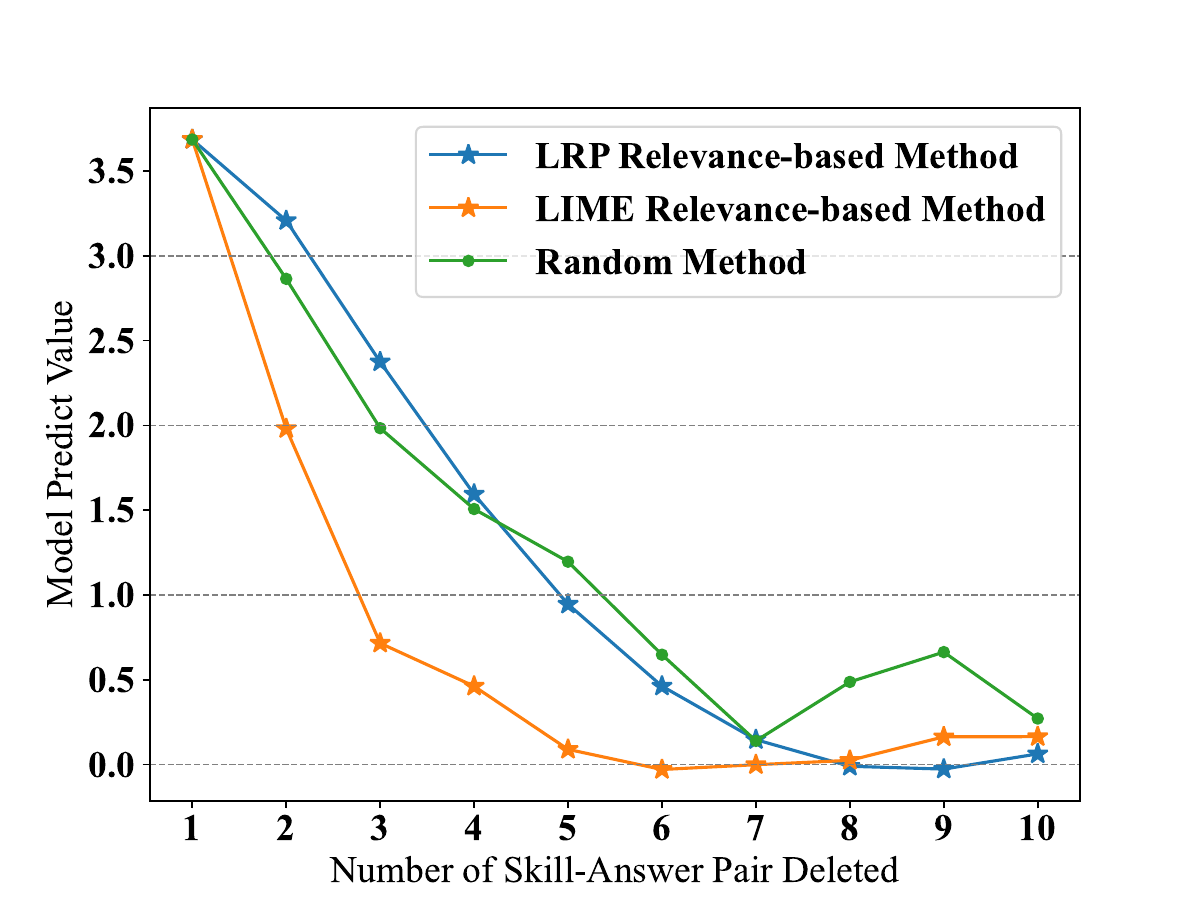}
    \caption{\textrm{Deletion experiment for LRP and LIME methods.}}
    \label{dete}
\end{figure*}

In Fig. \ref{compare}, Row  1 indicates the question ID, Row 2 indicates the correctness of the corresponding question, where 1 represents the correct answer, and 0 represents an incorrect answer, Row 3 is the model prediction, which indicates the probability of correct answers, and Rows 4-6 represent the eigenvalues calculated by LRP, LIME, and SHAP, respectively. The model correctly predicts position 24 for student 1, and the LIME method better handles the phenomenon of incorrect prediction at positions 14, 18, and 21. Similarly, when the model predicts the problem at position 24 with a low probability (0.56) for Student 2, all three methods can assign different eigenvalues to the irrelevant problem at position 23. Through the above two examples, it is found that the LRP method tends to transition the feature contribution of continuous problems smoothly, the LIME method distributes the feature contribution more evenly, and the SHAP method is more focused on sharing the features of current problems.

In the deletion experiment, as shown in Fig. 14, three lines representing different deletion strategies—random deletion (green), deletion based on the LIME method—for which the feature importance was calculated (yellow) and deletion based on the LRP method—for which the relevance was calculated (blue) are observed. The x-axis represents the deletion of 0 to 10 pairs of the original input data, and the y-axis represents the LSTM model’s predicted values. Before x=4, the LRP line is below the green line, the green line is below the LIME line, and the LIME line shows the steepest decrease. The difference between the green and LRP lines is approximately 0.5, and the difference between the green and LIME lines is approximately 1-2. After x=4, the LIME line continues to decline rapidly, reaching a plateau after x=6. The descent of the LRP line accelerates after x=4, decreases below the green line by approximately 0.5, and plateaus after x=8. The green line remains above the other lines, exhibiting fluctuations after x=7 but converging with the other lines at approximately x=10.

In conclusion, different deletion strategies significantly impact the model interpretability and robustness. The LIME method shows high sensitivity to small deletions but plateaus for larger counts. The LRP method performs better at capturing model changes with larger deletions. Random deletion shows relative robustness but may not capture complex model changes. The results emphasize the importance of choosing appropriate feature deletion strategies for interpreting model behavior. Further research could explore combining different interpretation methods for a comprehensive understanding of model behavior.

All three approaches explain deep knowledge tracing models, but which approach is closest to reality? The interpretability of models has become an urgent problem. The following sections briefly introduce the common evaluation methods used in KT and xAI and then explore how we can develop a standardized and reasonable interpretable evaluation system for educational models on knowledge tracing tasks. The goal is to improve the user’s understanding and trust in the education model and realize the wide application of intelligent education products in education.

\subsection{Common Evaluation Metrics for Knowledge Tracing}
The accuracy (ACC) and area under the curve (AUC) are the two main metrics commonly used to evaluate the performance of knowledge tracing models. The accuracy represents the proportion of correct prediction results among all the results. The AUC is the area under the ROC curve, and a lower coordinate axis indicates that the probability of a positive prediction is greater than that of a negative prediction. Therefore, the higher the AUC value is, the better the model being evaluated can achieve classification. However, it's important to note that while these metrics are instrumental in evaluating the model's predictive accuracy, they do not contribute to the evaluation of the model's interpretability.

\subsection{Evaluation Metrics for Explainable Knowledge Tracing}
Inspired by the classification of xAI evaluation methods \cite{vilone2021notions} in the previous section, in this section, the evaluation methods for explainable knowledge tracing are presented, summarised in Table. \ref{SEM} . As discussed above, “humans” are the main component of the whole educational loop. Therefore, to evaluate xKT, in this paper, models are evaluated from a subjective perspective, that is, focusing on evaluating xAI systems concerning the target audience and specific interpretability goals. Therefore, we divide the educational subjects into three categories in the task of knowledge tracing: KT educators, learners, and developers. Considering the different interpretability goals of the three types of stakeholders in the educational process, in this section, we elaborate on the interpretability of each category of subjects in the follow-up.

\begin{table*}[t]
    \caption{Summary of Evaluation Metrics for xKT.}
    \label{SEM}
    \resizebox{\textwidth}{!}{
    \small
    \begin{tblr}{
        colspec={Q[l,m,0.2\textwidth] Q[l,m,0.15\textwidth] Q[l,m,0.6\textwidth] Q[l,m,0.2\textwidth]},
        hspan=minimal,
    }
    \hline[1pt]
    User Category	       & Stakeholders	  & Evaluation Metrics	                      & References    \\   \hline
    Professional Users             & Developers  & Objective metric(stability, fidelity, sensibility, ect.)\newline Human-machine interaction \newline Multi-disciplines integration & \cite{vilone2021notions} \cite{alvarez2018towards}\newline \cite{kosch2023survey} \cite{ mosqueira2023human} \newline \cite{rapp2023processing} \cite{andrews2023role} \cite{kubiszyn2024educational}    \\ 
    \hline
    Non-professional Users & Educators, Learners                                  & Subjective metrics (interviews, questionnaires, scale analyses)\newline Social experiment &    \cite{10.1145/2939672.2939874} \cite{,10.1145/3397271.3401032} \newline \cite{alam2022contemplative}                       \\
    \hline[1pt]
    \end{tblr}
    }
\end{table*}

\subsubsection{From the Perspective of Professional Users}
From the perspective of KT developers, according to the xAI user classification standard in the previous section \cite{vilone2021notions}, developers are considered professional users. In general, developers are usually AI scientists and data engineers who design machine learning models and interpretable techniques for xAI systems. Compared with nonprofessional users, professional users are clearer about the operation mechanism of the model. How to provide reasonable and scientific explanations according to the different needs and abilities of different end users is a problem that professional users need to consider.

\textbf{Adopt objective metrics.} Developers should leverage quantitative evaluation methodologies within the realm of xAI for evaluating xKT processes. These methodologies include the deployment of standardized objective metrics, such as stability \cite{vilone2021notions}, fidelity \cite{vilone2021notions}, and sensibility \cite{alvarez2018towards}, to appraise the congruence in xKT’s interpretation of proximate or analogous data instances and the accuracy in approximating black-box model predictions. However, it is pertinent to acknowledge that the technical nature of these methods may render them less accessible to laypersons, potentially impeding their efficacy in broader evaluative contexts.

\textbf{Human-machine interaction.} We advocate for the integration of advanced human-machine interaction \cite{kosch2023survey, mosqueira2023human} technologies to facilitate a dynamic interaction loop between users and AI models. In this loop, the AI system should adapt its outputs based on feedback from the user, who may function as a teacher or learner, utilizing actions such as modifying data labels or assessing the validity of decisions made by the model. This approach fosters a more immersive evaluation of the model interpretability, striving to harmonize human intuition with artificial intelligence insights for more effective assessment. Essential to this process is the active engagement between developers and end users, as user-centric feedback is critical for the iterative refinement of explanatory mechanisms offered by the model.

\textbf{Multi-disciplines integration.} Finally, a collaborative approach with specialists in cognitive psychology or educational theory \cite{rapp2023processing} is recommended. Cognitive psychology principles, particularly mental model theories \cite{andrews2023role}, can be instrumental in conceptualizing a framework for understanding human-machine interaction and behavior. This understanding is crucial for an effective evaluation of AI interpretability. Furthermore, incorporating insights from educational measurement theories \cite{kubiszyn2024educational} enables developers to ascertain whether AI model predictions align with established cognitive learning patterns, thereby facilitating a more robust evaluation of the model interpretability.

\subsubsection{From the Perspective of Non-professional Users}
From the perspective of evaluating the interpretability of a model with teachers and learners as the subject, according to the xAI user classification standard in the previous section, teachers and learners are considered nonprofessional users, who do not understand the internal structure and operation mechanism of the model and consider an AI model a “black box”. Therefore, how to improve the model transparency and the user’s reliance is a problem that requires increased attention.

\textbf{Adopt subjective metrics.} Researchers can choose the subjective evaluation method in xAI. For example, interviews, questionnaires, and scale analyzes are used to evaluate the validity of explanations provided models and the satisfaction and trust of users so that nonprofessional users can understand and trust model decisions \cite{10.1145/2939672.2939874,10.1145/3397271.3401032}. Before this process, we should improve the AI literacy of teachers and learners in advance so that they can provide reasonable and scientific feedback. 

\textbf{Social experiment.} It is also possible to attempt to design a reasonable social experiment of intelligent education to evaluate the interpretability of a model \cite{alam2022contemplative}. In brief, researchers can recruit some stakeholders involved in knowledge tracing tasks and conduct small social experiments to empirically study the interpretability of a model. For example, participants can evaluate the interpretability of a model by reviewing the interpretation results.

Generally, in this section, we first analyze the limitations of the current evaluation metrics for knowledge tracing models. Next, inspired by the evaluation methods of xAI, we review the evaluation methods of the xKT model. Specifically, the evaluation methods proposed in this paper are human-centered and can be subdivided into methods for professional users (developers) and nonprofessional users (educators and learners). According to the characteristics of the two types of target users, corresponding evaluation methods have been proposed. The goal of this section is to provide some ideas for evaluating explainable knowledge tracing.

\section{Explainable Knowledge Tracing : Future Directions}\label{sub5}

xKT is emerging at the crossroads of xAI and educational analysis, driven by the need for effective, comprehensible, and ethically sound models in education. We will explore four key future directions that have been identified for their potential to substantially enhance xKT: Balancing model performance with interpretability to create sophisticated yet transparent algorithms, making advanced models accessible through user-friendly explainable methods, integrating causal inference to shift from prediction to understanding learning dynamics, and addressing ethical and privacy concerns in the data-driven educational era. These areas collectively aim to enhance xKT, aligning technical prowess with evolving educational and ethical demands.

\textbf{The Trade-Off Between Model Performance and Interpretability in Knowledge Tracing.}
In knowledge tracing, the critical future challenge is to balance model accuracy with interpretability. Achieving this balance requires creating algorithms that are both precise in prediction and intuitive in understanding. Future research is expected to focus on refining deep learning architectures to simplify structures and integrate advanced attention mechanisms, aiming to balance high performance with better interpretability \cite{lee2022monacobert}. Additionally, a growing trend is the integration of post hoc interpretability tools such as LIME \cite{mandalapu2021we} and SHAP \cite{valero2023shap,kim2021student,wang2022generic}, which offer clearer explanations for complex model decisions and uncover the underlying drivers of behavior. Moreover, complex, accurate models are likely to be blended with simpler, more interpretable models using advanced ensemble learning techniques \cite{shah2020explainable,sun2022ensemble}. This blend aims to improve the prediction accuracy while maintaining decision transparency, promoting enhanced educational quality and personalized learning in knowledge tracing applications.

\textbf{User-friendly Explainable Methods.}
As knowledge tracing technology evolves, user-friendly explainable methods have emerged as a core issue \cite{Lombrozo2016Explanatory}. Future research should focus on designing explainability mechanisms that are transparent not only to data scientists but also accessible and friendly to educators and learners. This level of explainability requires models to produce predictions that are easy to understand, in addition to clear logic and reasoning processes. Leveraging advanced natural language processing technology, models can generate detailed and comprehensible explanations, clearly articulating the logic behind their predictions. Furthermore, dynamic and interactive visualization technologies \cite{williamson2021effects} play a crucial role in intuitively presenting learners’ knowledge states and learning paths, significantly enhancing educators’ and learners’ understanding and acceptance of model feedback. Additionally, when designing these models, the intuitiveness of the user interface should be considered \cite{ghai2021explainable,conati2021toward}, enabling nonexperts to easily interpret and utilize the model outputs.

\textbf{Integrating Causal Inference into Knowledge Tracing Models.}
Integrating causal inference into knowledge tracing models is a vital direction for future research. This approach aims to uncover the actual causal relationships within learning processes, moving beyond the limitations of correlation-based analysis prevalent in many machine learning models \cite{li2023genetic}. By applying techniques such as counterfactual reasoning \cite{cornacchia2023auditing}, researchers can explore various hypothetical scenarios, and such alternative learning strategies might lead to diverse learning outcomes. This method enables a more profound understanding of the direct impact of specific learning activities on educational results. Such in-depth causal analysis not only improves the scientific rigor and accuracy of knowledge tracing models but also offers valuable insights for the development of effective and personalized educational interventions \cite{li2023genetic,zhu2023stable}. Consequently, knowledge tracing technology has advanced from simply predicting outcomes to providing actionable insights for enhancing educational practices.

\textbf{Ethical and Privacy Considerations in Model Explainability Research.}
In the realm of knowledge tracing, as efforts intensify to enhance model explainability, parallel emphasis must be placed on ensuring ethical and privacy considerations \cite{adams2023ethical}. Research should aim to design explainable models that not only provide transparent and understandable predictions but also rigorously protect user data and maintain ethical integrity. This research will involve developing techniques that balance the need for clarity in how models process and interpret personal data with robust measures to secure data privacy. Approaches such as differential privacy \cite{yang2023model,vasa2023deep}, which anonymizes data to prevent the identification of individuals, can be integrated with explainable Al frameworks. These approaches ensure that while models remain interpretable and that their decisions are transparent to users, they also adhere to strict privacy and ethical guidelines. Such research would necessitate a nuanced approach where explainability does not compromise privacy and ethical standards guide the transparency of the models.

\section{Conclusion}
This survey is the first to provide a comprehensive survey of explainable knowledge about multiple dimensions, including concepts, methods, and evaluations. Specifically, according to the xAI classification criteria for the complexity of explainable object models, we classify the related models of explainable knowledge training into two categories: 1) transparent models and 2) black box models. Then, representative explainable methods are reviewed in three stages: ante-hoc stage, post-hoc stage, and other dimensions.  Additionally, includes an investigation into the applications of explainable knowledge tracing. Furthermore, to fill the gap in the evaluation methods of explainable knowledge tracing, we consider evaluation methods from the perspective of education stakeholders. Finally, future research directions for explainable knowledge tracing are explored. The field of explainable knowledge tracing is booming, and we aim to draw researchers’ attention to the interpretability of algorithms, improve the transparency and reliability of algorithms, and provide a foundation and insight for researchers who are interested in interpretable knowledge tracing.

\section*{Declarations}

\begin{itemize}

\item Competing interests. 

This work was supported by the National Natural Science Foundation of China (Grant number [62207013] and [6210020445]). 
\item Availability of data and access.

All data generated or analysed during this study are included in this published article [\cite{feng2009addressing}].
\item Authors' contributions.

Yanhong Bai: Conception and design of the survey and writing the original draft.  Jiabao Zhao:  Conception and design of the survey, writing, and review. Tingjiang Wei: Writing and performing experiments. Qing Cai: Supervision and review. Liang He: Project administration, Resources.

\end{itemize}


\bibliography{sn-bibliography}

\end{document}